\documentclass[manuscript,screen]{acmart}

\usepackage{multirow}
\usepackage{tabularx}
\usepackage{array}
\usepackage{booktabs}
\usepackage{amsmath}
\usepackage{makecell}
\usepackage{pifont}

\newcommand{\cmark}{\ding{51}}
\newcommand{\xmark}{\ding{55}}

\AtBeginDocument{%
  }

\setcopyright{acmlicensed}
\copyrightyear{2018}
\acmYear{2018}
\acmDOI{XXXXXXX.XXXXXXX}

\acmConference[Conference acronym 'XX]{Make sure to enter the correct
  conference title from your rights confirmation emai}{June 03--05,
  2018}{Woodstock, NY}
\acmISBN{978-1-4503-XXXX-X/18/06}

\begin{document}

\title{A Survey on Knowledge-Oriented Retrieval-Augmented Generation}

\author{Mingyue Cheng}
\author{Yucong Luo}
\author{Jie Ouyang}
\author{Qi Liu}
\authornote{Corresponding author.}
\email{qiliuql@ustc.edu.cn}
\author{Huijie Liu}
\author{Li Li}
\author{Shuo Yu}
\author{Bohou Zhang}
\author{Jiawei Cao}
\author{Jie Ma}
\author{Daoyu Wang}
\author{Enhong Chen}
\email{cheneh@ustc.edu.cn}
\affiliation{%
  \institution{State Key Laboratory of Cognitive Intelligence, University of Science and Technology of China}
  \city{Hefei}
  \state{Anhui}
  \country{China}
}

\renewcommand{\shortauthors}{Cheng et al.}

\begin{abstract}
Retrieval-Augmented Generation (RAG) has gained significant attention in recent years for its potential to enhance natural language understanding and generation by combining large-scale retrieval systems with generative models. RAG leverages external knowledge sources, such as documents, databases, or structured data, to improve model performance and generate more accurate and contextually relevant outputs. This survey aims to provide a comprehensive overview of RAG by examining its fundamental components, including retrieval mechanisms, generation processes, and the integration between the two.  We discuss the key characteristics of RAG, such as its ability to augment generative models with dynamic external knowledge, and the challenges associated with aligning retrieved information with generative objectives. We also present a taxonomy that categorizes RAG methods, ranging from basic retrieval-augmented approaches to more advanced models incorporating multimodal data and reasoning capabilities. Additionally, we review the evaluation benchmarks and datasets commonly used to assess RAG systems, along with a detailed exploration of its applications in fields such as question answering, summarization, and information retrieval. Finally, we highlight emerging research directions and opportunities for improving RAG systems, such as enhanced retrieval efficiency, model interpretability, and domain-specific adaptations. This paper concludes by outlining the prospects for RAG in addressing real-world challenges and its potential to drive further advancements in natural language processing\footnote{https://github.com/USTCAGI/Awesome-Papers-Retrieval-Augmented-Generation}.
\end{abstract}


\received{20 February 2025}

\maketitle

\section{Introduction}
Retrieval-Augmented Generation (RAG)\cite{NEURIPS2020_6b493230} has emerged as a key approach that integrates information retrieval with generative models to enhance natural language processing tasks. By leveraging external knowledge sources, RAG systems can generate more accurate and contextually relevant outputs, addressing complex challenges in areas like question answering\cite{kwiatkowski2019natural}, summarization\cite{hayashi2021wikiasp}, and open-domain dialogue. In recent years, a variety of RAG methods have been proposed, ranging from basic retrieval-augmented models to more advanced architectures incorporating multi-hop\cite{press2022measuring} reasoning and memory-augmented techniques\cite{fevry2020entities}. These developments have highlighted the potential of RAG to improve the performance of NLP systems by combining retrieval and generation in a unified framework.

RAG models augment traditional language models by incorporating external knowledge sources, such as documents, databases, or structured data\cite{huang2024moremakingsmallerlanguage, jiang2023structgptgeneralframeworklarge}, during the generation process. Unlike conventional models that rely solely on pre-trained parameters, RAG systems dynamically retrieve relevant information at generation time, allowing them to produce more informed and contextually accurate outputs. This approach addresses key limitations of traditional language models, such as their inability to access real-time or domain-specific knowledge, and mitigates the challenge of handling out-of-vocabulary or rare entities. For example, in question answering tasks\cite{fan2019eli5,qin2023webcpm}, RAG models retrieve relevant passages from large corpora to generate more precise and informative answers, while in summarization\cite{hayashi2021wikiasp,narayan2018don}, they leverage external documents to provide richer and more comprehensive summaries. Early successes in RAG have demonstrated significant improvements in a range of NLP applications, including open-domain question answering, where RAG systems have outperformed traditional generative models by incorporating relevant external context, and document-based summarization, where they produce summaries that better reflect the nuances of the source material.

At the core of RAG lies a knowledge-centric approach, which places external knowledge as a key factor in improving language generation. By incorporating relevant, real-time, and structured information, RAG models can significantly enhance their ability to generate contextually accurate and factually grounded content. This knowledge-centric perspective addresses one of the critical limitations of traditional language models, which are constrained by their training data and lack access to dynamic or domain-specific knowledge. The integration of external knowledge allows RAG models to not only retrieve and incorporate relevant details but also reason over multiple pieces of information, resulting in more nuanced and informed outputs. This shift towards knowledge augmentation enables models to perform more complex tasks, such as handling specialized topics\cite{wellawatte2024chemlit,zhang2023enhancing,wiratunga2024cbr}, improving response relevance in dialogue systems, and generating high-quality summaries\cite{hayashi2021wikiasp,narayan2018don} that reflect the true essence of source materials. By managing the lifecycle of external knowledge, from retrieval and integration to reasoning and generation, RAG opens new possibilities for applications that demand a high level of accuracy and contextual awareness.

Despite the promising advancements in RAG, several challenges remain that hinder the full potential of these models. One of the primary issues is knowledge selection, where the model must effectively identify the most relevant pieces of information from vast external sources. This task is particularly challenging given the large, noisy, and diverse nature of real-world knowledge corpora. Another critical challenge is knowledge retrieval, which involves retrieving the right information at generation time while balancing efficiency and relevance. Current retrieval techniques often struggle with accurately matching queries to knowledge sources, especially in highly specialized or ambiguous contexts. Additionally, knowledge-enhanced in-context reasoning presents a significant hurdle, as models must not only retrieve relevant knowledge but also effectively integrate and reason over it in a coherent and contextually appropriate manner. These challenges highlight the need for a deeper understanding and improved methodologies in the RAG space, making a comprehensive survey necessary to address both the current limitations and future opportunities in this rapidly evolving field.

The primary objective of this survey is to provide a comprehensive overview of RAG from a knowledge-centric perspective. Unlike existing surveys that often categorize RAG methods by model architecture or task, our unique contribution lies in structuring the analysis around the \textit{lifecycle of knowledge} within RAG systems. We examine RAG as a multi-stage process of identifying, retrieving, integrating, and reasoning with external information. This knowledge-centric viewpoint provides a unified framework that shifts the focus from \textit{what} models are used to \textit{how} knowledge is processed and utilized, offering a clearer understanding of the trade-offs between different techniques. We will cover key models and the fundamental approaches used in retrieval and generation, offering insights into how these methods address the unique challenges at each stage of the knowledge lifecycle. Additionally, we seek to highlight emerging trends and identify gaps in the existing research, particularly areas that require further exploration, such as multimodal knowledge integration\cite{pmlr-v139-brock21a,radford2021learningtransferablevisualmodels,NEURIPS2020_92d1e1eb} and domain-specific applications\cite{yu2024large,setty2024improving,modran2024llm,liu2024application}. As the field continues to evolve rapidly, this survey will serve as both a foundational resource and a guide to future research, offering insights into ongoing challenges and opportunities for innovation.

This survey is organized as follows to provide a structured and comprehensive exploration of RAG through our knowledge-centric lens. Section 2 reviews related work and the origins of RAG, while Section 3 presents an overview of the RAG pipeline and its core components. Section 4 describes the fundamental objectives and challenges of RAG. Section 5 introduces our taxonomy, followed by basic and advanced RAG approaches in Sections 6 and 7, respectively. Section 8 examines evaluation strategies, benchmarks, and metrics for RAG systems \cite{saadfalcon2024aresautomatedevaluationframework,es2023ragasautomatedevaluationretrieval,petroni2021kiltbenchmarkknowledgeintensive}. Section 9 discusses downstream tasks and real-world applications, and Section 10 outlines prospects and future research directions. Finally, Section 11 concludes the survey.

\begin{figure}
    \centering
    \includegraphics[width=1\linewidth]{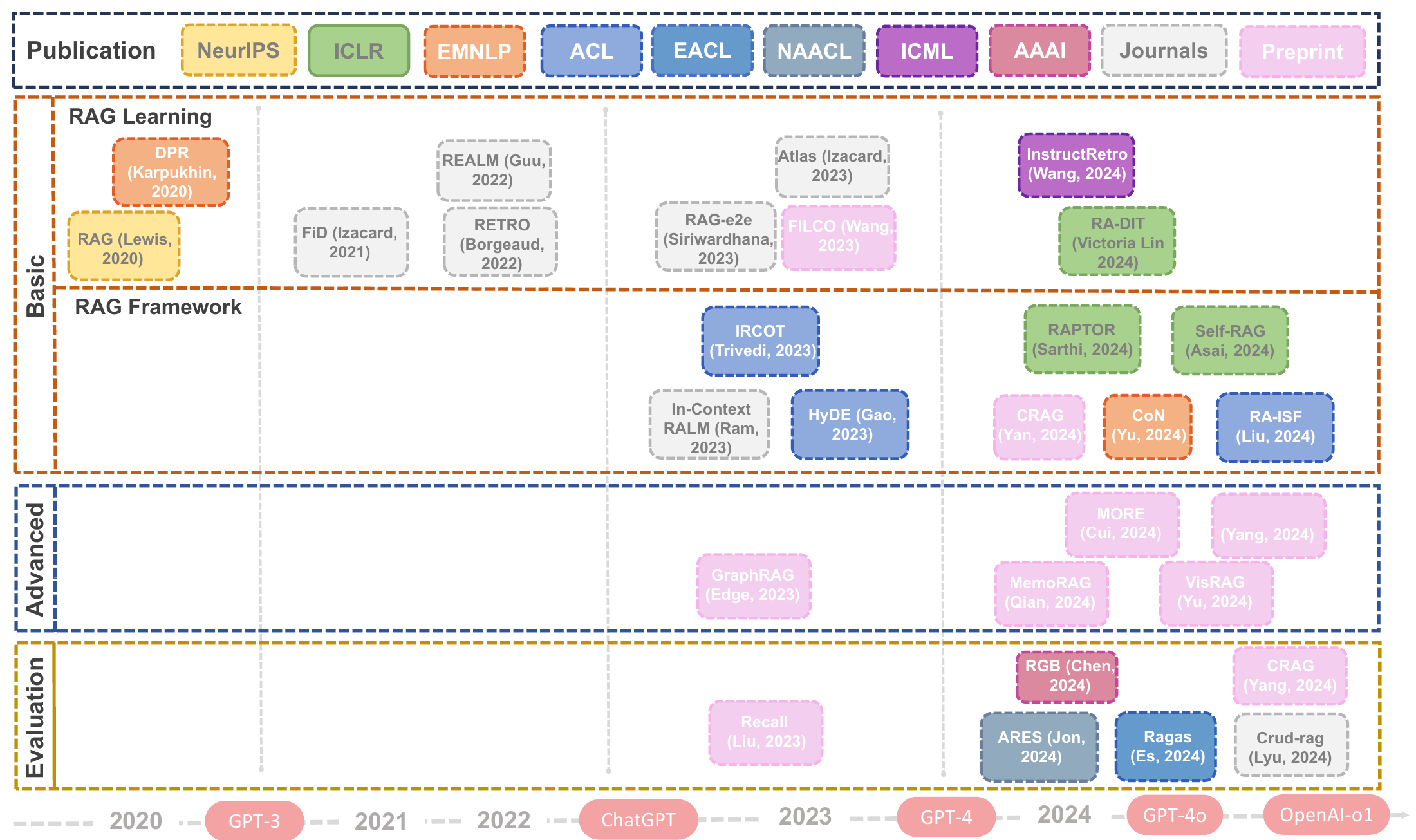}
    \caption{A Framework for Organizing RAG Works. The timeline spans from 2020 to the present, categorizing RAG-related research into three main areas: Basic (including RAG Learning and RAG Framework), Advanced, and Evaluation. Key milestones in language models (GPT-3, ChatGPT, GPT-4) are marked along the timeline. }
    \label{fig:framework}
\end{figure}
\section{Related Work}

This section provides essential background for understanding RAG. We begin by tracing the origin of RAG, exploring its evolution from its formal introduction and subsequent development.

\subsection{The Origin of RAG}

As language models evolved, researchers began exploring retrieval mechanisms to enhance text generation \cite{weston-etal-2018-retrieve,dinan2019wizardwikipediaknowledgepoweredconversational}. In 2020, Facebook formally introduced the concept of Retrieval-Augmented Generation (RAG) and successfully applied it to knowledge-intensive tasks \cite{lewis2020retrieval}, demonstrating that integrating external knowledge through retrieval substantially improves performance in question answering and text understanding. Concurrently, Google’s REALM \cite{guu2020retrieval} adopted a similar retrieval-augmented paradigm by incorporating a latent knowledge retriever during pre-training, achieving state-of-the-art results in open-domain question answering. Since then, RAG has attracted increasing research attention for its distinctive advantages over traditional generative models. The synergy between retrieval and generation enables models to dynamically access relevant external information during complex reasoning, thereby enhancing both the informativeness and factual accuracy of generated outputs. This paradigm marks a fundamental shift in natural language processing, where models no longer rely solely on internal parameters but effectively leverage vast external knowledge bases. As illustrated in Figure~\ref{fig:framework}, while early studies gradually explored various aspects of RAG since 2020, the field has witnessed exponential growth following the release of ChatGPT in late 2022, with numerous works emerging to further empower LLMs through retrieval-augmented approaches, continuously shaping the landscape of modern NLP.

\subsection{Related Surveys}
\begin{table}[t]
    \caption{Comparison of RAG surveys. \textbf{LLM}: whether the survey discusses RAG in the context of LLMs; \textbf{Multimodal}: whether it covers multimodal RAG; \textbf{Graph}: whether it discusses graph-structured information in RAG; \textbf{Advanced}: the coverage of advanced RAG techniques; \textbf{Evaluation}: whether it addresses evaluation methods; \textbf{Knowledge}: whether it takes a knowledge-centric perspective.}
    \label{tab:comparison}
    \centering
    \setlength{\tabcolsep}{6pt}
    \begin{tabular}{@{}lccccccccc@{}}
    \toprule
    & \textbf{Year} & \textbf{LLM} & \textbf{Multimodal} & \textbf{Graph} & \textbf{Advanced} & \textbf{Evaluation} & \textbf{Knowledge} \\ \midrule
    \citet{li2022surveyretrievalaugmentedtextgeneration} & 2022 & \xmark & \xmark & \xmark & \xmark & \xmark & \xmark \\
    \citet{gao2024retrievalaugmentedgenerationlargelanguage} & 2024 & \cmark & \xmark & \xmark & Training & \xmark & \xmark \\
    \citet{fan2024surveyragmeetingllms} & 2024 & \cmark & \xmark & \xmark & Training,Adaptive & \xmark & \xmark \\
    \citet{huang2024surveyretrievalaugmentedtextgeneration} & 2024 & \cmark & \xmark & \xmark & Training,Adaptive & \xmark & \xmark \\
    \citet{wu2024retrievalaugmentedgenerationnaturallanguage} & 2024 & \cmark & \xmark & \xmark & Training,Adaptive & \xmark & \xmark \\
    \citet{zhao2023retrievingmultimodalinformationaugmented} & 2023 & \cmark & \cmark & \xmark & \xmark & \xmark & \xmark \\
    \citet{zhao2024retrievalaugmentedgenerationaigeneratedcontent} & 2024 & \cmark & \cmark & \xmark & \xmark & \xmark & \xmark \\
    \citet{peng2024graphretrievalaugmentedgenerationsurvey}  & 2024 & \cmark & \xmark & \cmark & \xmark & \xmark & \xmark \\
    \citet{yu2024evaluationretrievalaugmentedgenerationsurvey} & 2024 & \cmark & \xmark & \xmark & \xmark & \cmark & \xmark \\
    (Ours) & 2025 & \cmark & \cmark & \cmark & All & \cmark & \cmark \\
    \bottomrule
    \end{tabular}
\end{table}
With the ongoing advancements in the field of generative AI, specifically in RAG, numerous surveys have emerged. However, these surveys often focus on specific aspects of the field. They either concentrate solely on a single foundation of RAG or provide a concise overview of enhancement methods for RAG in limited scenarios. Most existing works emphasize text-related RAG tasks supported by large language models without delving into other modalities. The survey by Li et al.\cite{li2022surveyretrievalaugmentedtextgeneration} offers a fundamental overview of RAG and discusses specific applications within the realm of text generation tasks. Recent surveys by Gao et al. and Fan et al. \cite{gao2024retrievalaugmentedgenerationlargelanguage,fan2024surveyragmeetingllms,huang2024surveyretrievalaugmentedtextgeneration} explore RAG in the context of large language models, with a specific emphasis on query-oriented RAG enhancement methods. Wu et al. \cite{wu2024retrievalaugmentedgenerationnaturallanguage} delve into key RAG technologies in information retrieval while also introducing broad applications in natural language processing tasks. Several other works investigate a more general approach to RAG. Recent works by Zhao et al. \cite{zhao2023retrievingmultimodalinformationaugmented,zhao2024retrievalaugmentedgenerationaigeneratedcontent} extend RAG to multimodal contexts, exploring its technologies and applications in broader AI-generated content (AIGC) scenarios. Another work by Peng et al. \cite{peng2024graphretrievalaugmentedgenerationsurvey} examines how graph-structured information can aid more precise and comprehensive retrieval in RAG, enhancing relational knowledge acquisition and the generation of context-aware responses. In addition to concentrating on research pertaining to RAG technology, recent scholarship has increasingly directed its focus towards the assessment of RAG systems. A survey by Yu et al.\cite{yu2024evaluationretrievalaugmentedgenerationsurvey} addresses the evaluation challenges of RAG systems, providing an integrated evaluation framework and examining existing benchmarks along with their limitations. Despite covering multiple aspects of RAG, existing surveys still lack a comprehensive review that includes discussions on RAG foundations, enhancement methods, and their applications across different domains. Furthermore, current surveys generally overlook the essence of RAG, namely knowledge utilization. This paper aims to fill this gap through a systematic survey, exploring RAG with a knowledge-centric approach.

\section{Overview of Retrieval-Augmented Generation}


RAG has emerged as a powerful paradigm by integrating retrieval mechanisms into the generation pipeline, RAG addresses the limitations of traditional sequence-to-sequence frameworks, particularly in scenarios requiring extensive domain knowledge. This section presents a comprehensive overview of RAG, discussing its core components: \textbf{retrieval} from external knowledge sources, \textbf{generation} with both internal and external knowledge, and the critical process of \textbf{knowledge integration} that bridges these elements.

\begin{figure}
    \centering
    \includegraphics[width=0.9\linewidth]{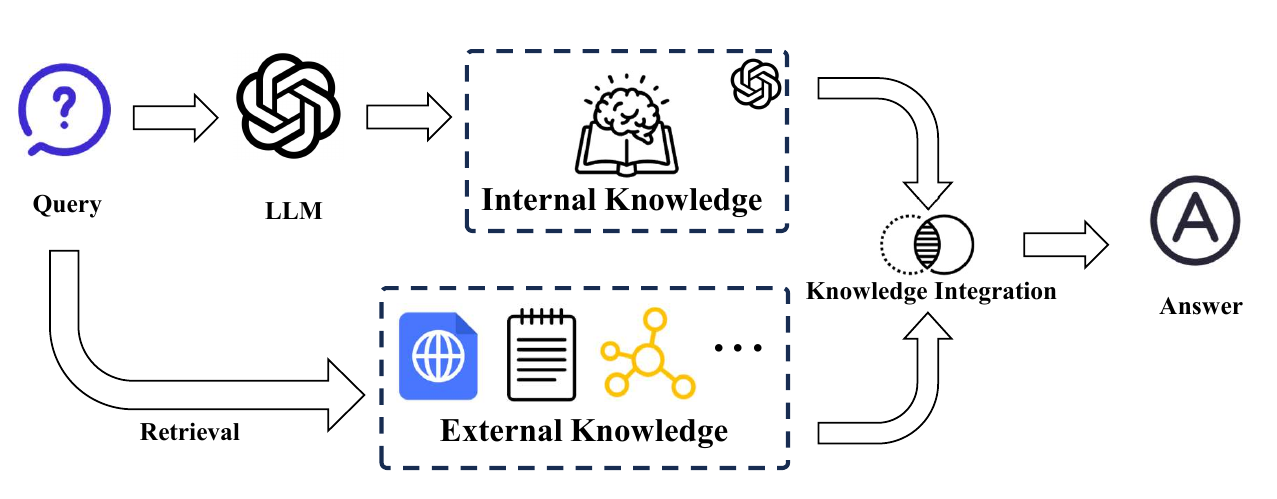}
    \caption{Overview of RAG. The framework consists of three main components: (1) A query is processed by an LLM with its internal knowledge, (2) External knowledge is retrieved based on the query, and (3) Knowledge integration combines both internal and external knowledge to generate the final answer.}
    \label{fig:overview}
\end{figure}

\subsection{Problem Formulation}

Most generation tasks can be conceptualized as a transformation from an input sequence $\boldsymbol{x}$ to an output sequence $\boldsymbol{y}: \boldsymbol{y} = f(\boldsymbol{x})$. However, this basic framework may be limited by insufficient information or complex contextual requirements. RAG addresses this limitation by introducing a retrieval component that enriches the input with relevant external knowledge. Specifically, a retrieval function $g$ extracts relevant information $\boldsymbol{z}$ from external knowledge repositories: $\boldsymbol{z} = g(\boldsymbol{x})$. The generation function $f$ then combines the input sequence $\boldsymbol{x}$ with the retrieved context $\boldsymbol{z}$ to produce the output: $\boldsymbol{y} = f(\boldsymbol{x}, \boldsymbol{z}) = f(\boldsymbol{x}, g(\boldsymbol{x}))$. This enhanced framework leverages external knowledge to improve generation quality, particularly beneficial for tasks requiring complex contextual understanding like machine translation and visual question answering.

\subsection{Retrieval}

The retrieval process in RAG aims to provide useful knowledge from external sources. Such knowledge may take multiple modalities, including text, image, video, audio, code, table, etc. Correspondingly, the storage formats are diverse, spanning from simple text files to complex databases and search engines. Additionally, the structure of this knowledge is heterogeneous; it may exist as unstructured plain text, as semi-structured html data or as structured graph data. Any process of obtaining relevant information from external knowledge bases can be considered as retrieval, irrespective of the modality, storage format, or structure of the knowledge.

\subsection{Generation}

In the generation process, generative models incorporate both internal and external knowledge to convert inputs into coherent and pertinent outputs. Regardless of the data format or task, the essence of the generation phase lies in "Denoising" \cite{asai2023self,wei2024instructrag} and "Reasoning" \cite{wang2024rat,trivedi-etal-2023-interleaving}.  Denoising focuses on filtering out irrelevant or contradictory information from retrieved knowledge, ensuring that only reliable and relevant information influences the generation process. Meanwhile, reasoning enables the model to effectively synthesize information from multiple sources, draw logical connections, and produce well-founded outputs. Through effective "Denoising" of retrieved information and sophisticated "Reasoning" over multiple knowledge sources, generative models can handle complex tasks while maintaining output accuracy and coherence.

\subsection{Knowledge Integration}

Knowledge integration, often referred to as \textit{augmentation} in the context of RAG, is a critical process that integrates internal knowledge from LLMs with retrieved external knowledge. This neural-symbolic integration can be achieved through three main approaches. Input-Layer Integration \cite{izacard-grave-2021-leveraging,ram-etal-2023-context} concatenates retrieved documents directly with the original query before feeding into the model, allowing simultaneous processing of both query and external knowledge. Output-Layer Integration \cite{khandelwal2020generalizationmemorizationnearestneighbor} incorporates retrieved knowledge at the logits level to calibrate the model's final predictions, particularly effective for improving output accuracy. Intermediate-Layer Integration \cite{borgeaud2022improving} integrates external knowledge into the model's hidden states during generation, enabling more nuanced interaction between internal and external knowledge representations. Each integration strategy offers distinct advantages and can be selected based on specific task requirements and computational constraints.

\section{Fundamentals and Objectives of RAG}
\begin{figure*}
    \centering
    \includegraphics[width=1\linewidth]{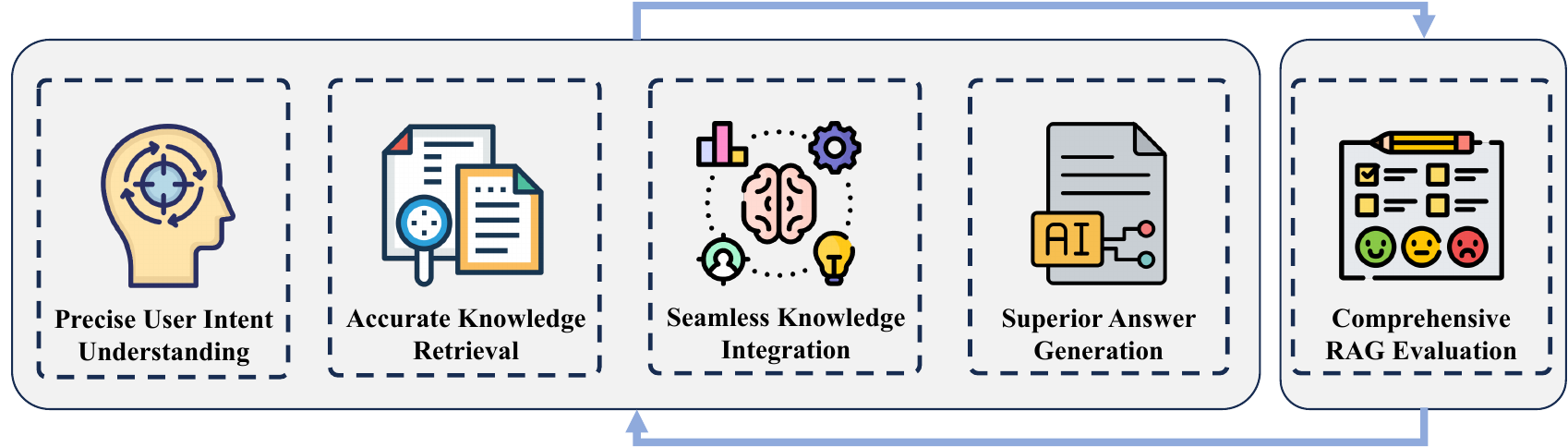}
    \caption{Fundamentals and objectives of RAG, including user intent understanding, knowledge retrieval, knowledge integration, answer generation, and RAG evaluation.}
    \label{fig:fundamentals}
\end{figure*}

RAG models have emerged as a powerful paradigm that combines the strengths of information retrieval and answer generation. By leveraging external knowledge sources, RAG models improve the quality and relevance of generated content, making them effective for various applications such as question answering, summarization, and conversational agents. However, the integration of retrieval and generation components introduces a complex set of characteristics and challenges that must be addressed to fully realize the potential of RAG systems. This section explores the key aspects of RAG, examining the main components of user intent understanding, knowledge retrieval, knowledge integration, answer generation, and evaluation metrics, while providing deeper insights into each area, as shown in Figure \ref{fig:fundamentals}.

\subsection{Precise User Intent Understanding}
Accurately understanding user intent \cite{luo2024unlocking,zhang2023towards,chen2012understanding} is essential for RAG models to generate responses that are semantically relevant and contextually appropriate. However, user queries are often ambiguous, implicit, or influenced by personalized preferences, making intent interpretation challenging. Subtle linguistic cues, such as vague expressions, informal phrasing, or culturally dependent references, further obscure user intent, hindering the model’s ability to extract key information. Consequently, effective intent understanding demands approaches that move beyond surface-level keyword matching toward deeper semantic and contextual analysis.

To overcome these challenges, RAG models must implement advanced intent understanding strategies that incorporate contextual understanding and semantic analysis. Techniques such as text modeling \cite{alghamdi2015survey}, query rewriting \cite{ma2023query,10.1145/3488560.3498516,qian-dou-2022-explicit}, and intent clustering \cite{10.1145/3488560.3498443} can enhance the LLM's ability to interpret user needs, leading to more precise retrieval and generation. By continuously refining these approaches, RAG models can better align with evolving user expectations and domain-specific requirements. Furthermore, accurate intent understanding reduces the risk of irrelevant or vague responses, improving user satisfaction. This, in turn, strengthens both retrieval and generation processes while also providing a foundation for delivering adaptable, high-quality responses across diverse and complex real-world scenarios.

\subsection{Accurate Knowledge Retrieval}
The knowledge retrieval phase in RAG models plays a pivotal role in supplying relevant context for generation, directly impacting the quality of the final output. High retrieval accuracy ensures access to the most pertinent knowledge, while efficiency prevents latency in real-time applications. A major challenge lies in managing large-scale datasets, which necessitate advanced indexing strategies such as inverted indices, approximate nearest neighbor (ANN) search \cite{arya1998optimal}, and scalable vector representations \cite{gormley2015elasticsearch} to maintain rapid access. Moreover, the heterogeneity and unstructured nature of data sources require robust preprocessing and normalization to ensure consistency and relevance. To further improve retrieval quality, TEKM \cite{10.1145/3442381.3449943} integrates semantic similarity, knowledge relevance, and topical relatedness, enhancing traditional retrieval frameworks through topic-aware embeddings and fine-grained kernel-based matching.

Balancing precision and recall is another critical aspect. High precision minimizes the retrieval of irrelevant contents, thereby reducing noise in the generative process. Conversely, high recall ensures comprehensive coverage of relevant information, which is essential for tasks requiring in-depth understanding. Achieving the above balance often involves fine-tuning retrieval algorithms and leveraging relevance feedback strategies. Moreover, the dynamic nature of information sources, where data is frequently updated or expanded, poses ongoing challenges for maintaining up-to-date and accurate retrieval strategies. Addressing these issues requires continuous adaptation and optimization of retrieval strategies to handle evolving datasets accurately and effectively.

\subsection{Seamless Knowledge Integration}
Integrating retrieved knowledge into LLM is a challenging task that demands seamless alignment between external knowledge and the LLM's internal knowledge \cite{grigoriou2017organizing}. Such integration is crucial for generating outputs that are not only semantically relevant but also coherent and contextually appropriate. One of the significant challenges is ensuring that the LLM can effectively interpret and utilize the retrieved information without introducing inconsistencies or factual inaccuracies, which requires strategies for merging diverse data types, such as text chunks, and structured data into a unified representation that LLM can process.

Furthermore, the integration phase must effectively manage potential conflicts between retrieved knowledge and the LLM’s pre-existing knowledge base. Inconsistencies or contradictions between newly retrieved information and the model’s internal knowledge can degrade response quality, necessitating advanced mechanisms for conflict resolution and knowledge validation to ensure harmonious integration. Temporal relevance also plays a critical role, especially in applications requiring current information. Thus, adaptive integration strategies that prioritize recent and contextually aligned knowledge are essential to preserve both the accuracy and timeliness of generated content.

\subsection{Superior Answer Generation}

Achieving superior answer generation in RAG models involves the complex task of synthesizing retrieved information with the LLM's natural language abilities to ensure outputs are accurate, relevant, and coherent. Such integration requires a refined alignment between external knowledge sources and the LLM's internal knowledge, reducing potential discrepancies that could lead to factual inconsistencies. Cross-referencing and validation strategies \cite{wei2024instructrag} play a vital role here, acting as safeguards that help verify the accuracy of information and prevent the spread of misinformation.  With these checks in place, the LLM can more reliably produce content that aligns with verified sources, enhancing the user's trust in the generated output.

Another key challenge lies in preserving fluency and naturalness in the generated text, particularly when synthesizing information from heterogeneous sources with differing styles and structures. The LLM must seamlessly integrate these inputs to produce outputs that are both coherent and contextually appropriate. Achieving such coherence requires advanced language modeling and strong contextual understanding to effectively reconcile diverse information. Moreover, the generation component must remain adaptable, dynamically adjusting to varying contexts and user intents. Whether providing concise answers, summarizing complex data, or engaging in dialogue, the model should flexibly tailor tone, style, and content to the communicative setting. This capability hinges on robust and versatile training strategies.

\subsection{Comprehensive RAG Evaluation}
Evaluating the performance of RAG presents a unique set of challenges due to the dual nature of retrieval and generation tasks. Traditional evaluation metrics, such as BLEU \cite{papineni2002bleu} and ROUGE \cite{lin2004rouge}, primarily focus on the quality of the generated text by comparing it to reference outputs. However, these metrics may not adequately capture the effectiveness of the retrieval component, which plays a crucial role in determining the relevance and accuracy of the generated content. To address this, comprehensive evaluation frameworks must integrate metrics that assess both retrieval accuracy, including precision, recall, and F1-score \cite{yacouby2020probabilistic}, and generation quality, including coherence, fluency, and factual accuracy.

Developing comprehensive evaluation metrics for RAG systems remains a complex task, requiring a balance between quantitative indicators and qualitative judgments to yield a holistic assessment of system performance. Aligning these metrics with real-world user satisfaction and practical utility further complicates the process, as subjective dimensions, such as perceived relevance, coherence, and helpfulness, are difficult to quantify yet crucial for measuring true effectiveness. Incorporating user studies, human evaluations, and adaptive, feedback-driven metrics can enhance both accuracy and contextual relevance. Moreover, ensuring metric robustness across diverse domains and applications is essential for maintaining generalizability. Finally, citation generation plays an important role in verifying the credibility and traceability of retrieved knowledge, reinforcing the reliability of RAG evaluation.

RAG systems include many different characteristics and challenges that span the entire process of intent
understanding, knowledge retrieval, knowledge integration, answer generation, and evaluation. Each component presents its own set of complexities, from handling large-scale and dynamic datasets in retrieval to ensuring seamless integration and high-quality generation of content. Addressing these challenges requires innovative solutions that leverage advanced algorithms, robust training methodologies, and comprehensive evaluation frameworks. Moreover, the interdependent nature of retrieval and generation tasks in RAG systems necessitates a complete approach to system design and optimization, ensuring that improvements in one area work together to improve overall performance. Continued research and development in these areas hold significant promise for advancing the capabilities of RAG models, enabling more intelligent, accurate, and contextually aware generative systems that can meet the evolving demands of diverse applications and industries.

\section{A Taxonomy of RAG Methods}

\begin{figure*}
    \centering
    \includegraphics[width=1.0\linewidth]{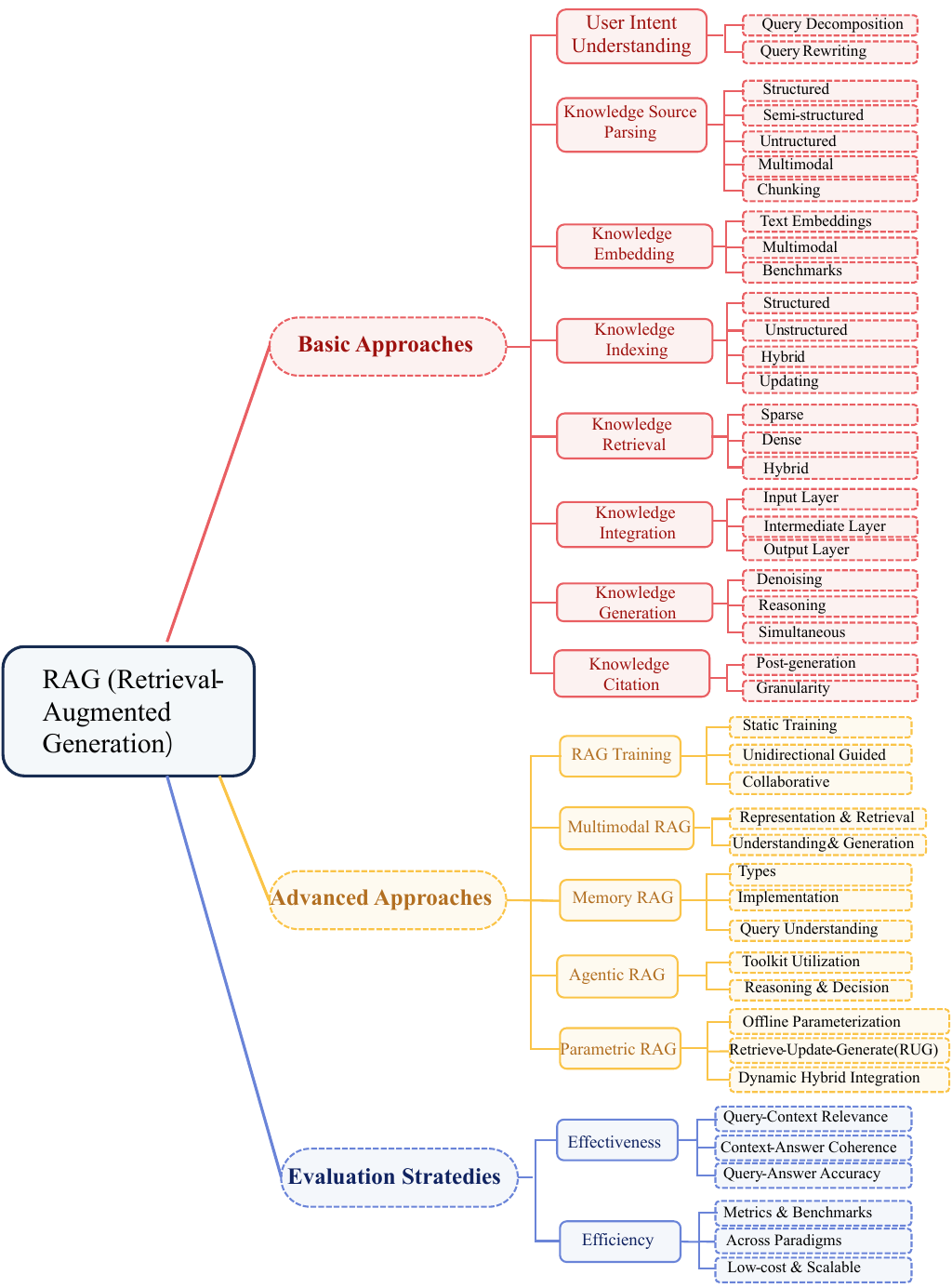}
    \caption{A systematic taxonomy of RAG. The field is organized into three core branches: (1) Basic Approaches, detailing the standard RAG pipeline; (2) Advanced Approaches, covering modern enhancements such as Agentic RAG, Multimodal RAG, and Parametric RAG; and (3) Evaluation Strategies, focusing on Effectiveness and Efficiency. }
    \label{fig:taxonomy}
\end{figure*}

As the field of RAG rapidly expands, a multitude of methods, architectures, and optimization techniques have emerged. To organize this complex and fast-evolving technological landscape and to provide a clear reference framework for researchers and practitioners, we propose a taxonomy of RAG methods.

As illustrated in Figure~\ref{fig:taxonomy}, our taxonomy organizes the RAG research landscape into three core branches: \textbf{basic approaches}, \textbf{advanced approaches}, and \textbf{evaluation strategies}. Basic approaches cover the core components of a RAG system and the complete workflow from receiving a query to generating a final, cited response, including user intent understanding, knowledge source and parsing, knowledge embedding, knowledge indexing, knowledge retrieval, knowledge integration, answer generation, and knowledge citation. Advanced approaches encompass techniques that extend and enhance the basic RAG pipeline, including RAG training, multimodal RAG, memory RAG, agentic RAG, and parametric RAG. Evaluation strategies assess RAG systems along two main dimensions: effectiveness, which measures generated-content quality through query-to-context relevance, context-to-answer coherence, and query-to-answer accuracy; and efficiency, which measures computational overhead and response performance in terms of latency, throughput, and resource utilization.

\section{Basic RAG Approaches}
\label{sec:basic}

\begin{figure*}
    \centering
    \includegraphics[width=1\linewidth]{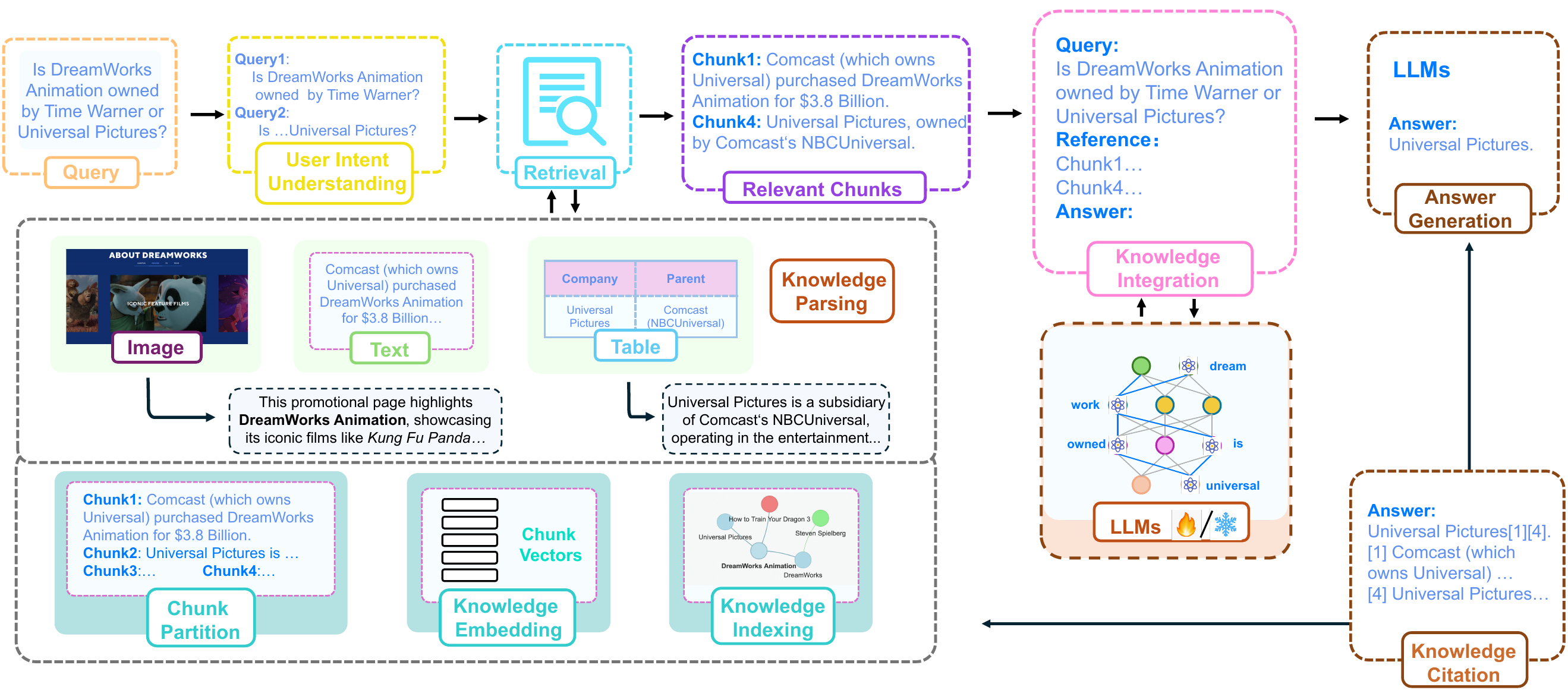}
    \caption{Basic RAG Approaches, including multi-source knowledge, embedding, indexing, retrieval and generation.}
    \label{fig:basic-final}
\end{figure*}

RAG models leverage external knowledge to enhance the generation process, enabling more accurate and contextual responses. The basic RAG approaches consists of several key steps: user intent understanding, knowledge source and parsing, knowledge embedding, knowledge indexing, knowledge retrieval, knowledge integration, answer generation, and knowledge citation. This section explains these core components, showing how RAG systems understand user queries, process different types of knowledge, convert information into vectors, build search indexes, retrieve relevant content, integrate knowledge with models, and generate answers with proper citations.

\subsection{User Intent Understanding}
High-quality queries are essential for retrieving valuable knowledge. Since users' intentions are often ambiguous, accurately interpreting user queries is crucial for enabling more effective and precise retrieval~\cite{cheng2021learning,cheng2022towards}. Currently, many studies focus on enhancing the understanding of user queries. This subsection primarily discusses two key methods for improving query quality: query decomposition and query rewriting.

\subsubsection{Query Decomposition}
Query decomposition has proven effective for enhancing the reasoning capabilities of language models, particularly in multi-step or compositional tasks. Methods such as least-to-most prompting \cite{zhou2022least} decompose complex problems into simpler subproblems, enabling strong generalization, as demonstrated by GPT-3’s over 99\% accuracy on SCAN with minimal examples. Self-ask \cite{press2022measuring} extends this idea by prompting models to generate and answer follow-up questions, narrowing the compositionality gap and improving multi-hop reasoning. Chain-of-verification (CoVe) \cite{dhuliawala2023chain} enhances reliability by verifying intermediate results through independent fact-checking, reducing hallucinations in complex queries. Similarly, search-in-the-chain (SearChain) \cite{xu2024search} integrates information retrieval within reasoning via a Chain-of-Query (CoQ), improving accuracy and traceability in knowledge-intensive tasks. More recently, Chain-of-Actions (CoA) \cite{pan2025chainofaction} unifies decomposition, real-time retrieval, and verification mechanisms to further strengthen reasoning robustness in LLMs.

\subsubsection{Query Rewriting}
Query rewriting has become a key technique for improving RAG performance by bridging semantic gaps and enhancing retrieval accuracy. Rewrite-Retrieve-Read (RRR) \cite{ma2023query} refines queries through an LLM-based rewriter trained with reinforcement learning, strengthening alignment between queries and target knowledge for open-domain QA and reasoning tasks. BEQUE \cite{peng2024large} targets long-tail queries in e-commerce via supervised fine-tuning, feedback, and contrastive learning, yielding substantial business gains. HyDE \cite{gao2022precise} adopts a zero-shot strategy where an LLM generates a hypothetical document for dense retrieval, outperforming traditional unsupervised methods. Step-Back Prompting \cite{zheng2023take} further enhances reasoning by abstracting high-level concepts from specific examples, improving multi-hop and knowledge-based tasks. Collectively, these approaches enhance both the precision and scalability of RAG across knowledge-intensive domains.

\subsection{Knowledge Source and Parsing}

The types of knowledge that RAG can utilize are diverse, providing extensive contextual information for LLMs. However, the parsing and efficient integration of these heterogeneous knowledge sources pose significant challenges. This subsection introduces the categories of knowledge employed by RAG, including structured, semi-structured, unstructured, and multimodal knowledge, along with their respective parsing and integration methodologies.


\subsubsection{Utilization of Structured Knowledge}
Knowledge graphs (KGs) are structured representations that encapsulate entities and their interrelations in a graph format, offering advantages for RAG. Their structured nature facilitates efficient querying and retrieval, while the semantic relationships they capture enable nuanced understanding and reasoning\cite{huang2024moremakingsmallerlanguage, jiang2023structgptgeneralframeworklarge}. Additionally, KGs integrate information from diverse sources, providing a unified knowledge base. However, integrating KGs into RAG systems presents challenges, including the complexity of navigating and extracting relevant subgraphs from extensive KGs, scalability issues as KGs expand, and aligning structured data with the unstructured data processing of language models.

Recent studies have advanced RAG through the integration of knowledge graphs (KGs) to enhance retrieval and reasoning. GRAG \cite{hu2024graggraphretrievalaugmentedgeneration} retrieves textual subgraphs across documents to improve retrieval efficiency, while KG-RAG \cite{sanmartin2024kgragbridginggapknowledge} employs the Chain of Explorations (CoE) algorithm for more effective KGQA. GNN-RAG \cite{mavromatis2024gnnraggraphneuralretrieval} leverages Graph Neural Networks to extract and process KG-based information, strengthening reasoning before interaction with LLMs. Constructed KGs from historical data also serve as external knowledge sources for RAG \cite{Xu_2024}, supporting more accurate generation. The SURGE framework \cite{kang2023knowledgegraphaugmentedlanguagemodels} further utilizes KG information for knowledge-grounded dialogues. Domain-specific systems such as SMART-SLIC \cite{barron2024domainspecificretrievalaugmentedgenerationusing}, KARE \cite{jiang2024reasoningenhancedhealthcarepredictionsknowledge}, ToG 2.0 \cite{ma2024thinkongraph20deepfaithful}, and KAG \cite{liang2024kagboostingllmsprofessional} demonstrate the effectiveness of KGs in specialized applications, enhancing accuracy and efficiency. Collectively, these advancements highlight the pivotal role of KGs in improving retrieval precision, reasoning depth, and domain adaptability in RAG systems \cite{10.1093/bioinformatics/btae560}.

\begin{figure*}
    \centering
    \includegraphics[width=1\linewidth]{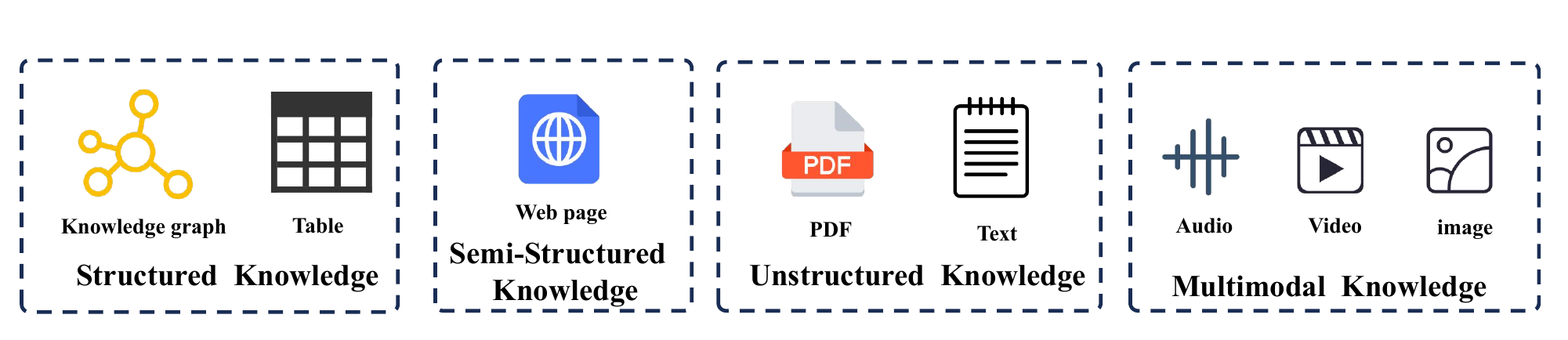}
    \caption{Diverse knowledge utilized by RAG, including structured, semi-structured, unstructured  and multimodal knowledge.}
    \label{fig:knowledge_source}
\end{figure*}

Tables convey rich information through the interplay of rows and columns, enhancing data density but posing challenges for language models to interpret\cite{wang2024chainoftableevolvingtablesreasoning}.   Tables often encapsulate large volumes of data succinctly, playing a crucial role in summarizing facts and quantitative data\cite{zhao2024tabpediacomprehensivevisualtable}. Additionally, the inherent organization of tables supports straightforward integration with various data processing tools and algorithms. However, integrating tabular data into RAG systems presents several challenges. One significant issue is the semantic understanding of table content, as tables may contain implicit relationships and context that are not immediately apparent. Another challenge is the diversity in table structures across different domains, which complicates the development of generalized models for table interpretation\cite{sui2024tablemeetsllmlarge}. Furthermore, translating natural language queries into structured queries that can effectively interact with databases requires sophisticated natural language processing techniques.

Recent research has tackled these challenges through the development of innovative methodologies. For instance, TableRAG\cite{chen2024tableragmilliontokentableunderstanding,deng2024tablestextsimagesevaluating}  utilizes query expansion along with schema and cell retrieval to accurately identify crucial information before passing it to language models. This approach allows for the efficient handling of large-scale tabular data, improving the accuracy and relevance of generated responses. Additionally, models like TAG\cite{chen2024tableragmilliontokentableunderstanding,deng2024tablestextsimagesevaluating} and extreme-RAG\cite{roychowdhury2024erattaextremeragtable} unify Text-to-SQL\cite{10.1145/3133887, 10.5555/1864519.1864543, yu-etal-2018-typesql} and RAG methods, broadening the range of user queries that can be addressed. Similarly, SLR\cite{10.1145/3609796} incorporates both internal and external structural signals, such as document segmentation and legal charge graphs, to better capture the semantics of lengthy, complex legal documents. These advancements underscore the potential of leveraging tabular data in RAG systems to enhance information retrieval and generation capabilities. By effectively integrating structured data, these systems can provide more precise and contextually relevant responses, thereby improving user experience and satisfaction.

\subsubsection{Extraction of Semi-Structured Knowledge}
Semi-structured data bridges the gap between structured and unstructured formats, combining organizational elements with flexible content. Common examples include JSON, XML, emails, and especially HTML, which structures text, images, and links through tags and attributes. While HTML’s hybrid nature enables rich information representation, its flexibility also introduces inconsistencies that complicate extraction and integration in RAG systems. Recent studies, such as HtmlRAG \cite{tan2024htmlraghtmlbetterplain}, explore methods for parsing HTML to enhance knowledge integration, yet open-source parsing tools remain essential for efficiency and adaptability. Libraries like Beautiful Soup \cite{richardson_beautiful_2024}, html5ever \cite{servo_html5ever_2024}, htmlparser2 \cite{muller_htmlparser2_2024}, MyHTML \cite{borisov_myhtml_2024}, and Fast HTML Parser \cite{fast_html_parser_2024} convert HTML into structured representations such as DOM trees, supporting efficient data manipulation and retrieval. These advancements in semi-structured data parsing provide critical infrastructure for integrating web-based content into RAG pipelines with improved accuracy and scalability.

\subsubsection{Parsing of Unstructured Knowledge}
Unstructured knowledge encompasses a broad array of data types that lack consistent structure, such as free-form text, and PDF documents. Unlike structured data, which adheres to predefined schemas, unstructured data varies widely in format and often contains complex content,  making direct retrieval and interpretation challenging. Among these unstructured formats, PDF documents are particularly prevalent in knowledge-intensive domains, including academic research, legal documents, and corporate reports. PDFs often house a wealth of information, with elements like text, tables, and embedded images, but their inherent structural variability complicates extraction and integration into RAG systems.

Parsing PDFs remains a complex challenge due to diverse layouts, fonts, and embedded structures. Converting them into machine-readable formats for RAG requires OCR, layout analysis, and interpretation of complex elements such as tables and formulas. Recent advances in vision-language models, such as Levenshtein OCR \cite{da2022levenshteinocr} and GTR \cite{he2021visualsemanticsallowtextual}, have improved OCR accuracy by combining visual and linguistic cues. Unified frameworks like OmniParser \cite{wan2024omniparserunifiedframeworktext} and Doc-GCN \cite{luo-etal-2022-doc}, as well as bidirectional OCR models like ABINet \cite{fang2021readlikehumansautonomous}, further enhance structural preservation and recognition performance.
In parallel, open-source tools for PDF-to-Markdown conversion have emerged as efficient solutions for unstructured data integration in RAG. Systems such as GPTPDF \cite{gptpdf}, Marker \cite{marker}, PDF-Extract-Kit \cite{PDF-Extract-Kit}, Zerox OCR \cite{Zerox}, MinerU \cite{MinerU}, and MarkItDown \cite{microsoft-markitdown} offer capabilities ranging from table and formula parsing to layout reconstruction and noise cleaning. These developments collectively mark a shift toward more accurate, multimodal, and structure-preserving pipelines for extracting and integrating PDF-based knowledge into RAG systems.

\subsubsection{Integration of Multimodal Knowledge}

Multimodal knowledge, including images, audio, and video, offers rich, complementary information that can significantly enhance RAG systems, especially for tasks requiring deep contextual understanding. Images provide spatial and visual details; audio contributes temporal and phonetic layers; and video combines both spatial and temporal dimensions, capturing motion and complex scenes. Traditional RAG systems, primarily designed for text-based data, often struggle to process and retrieve information from these modalities, leading to incomplete or less nuanced responses when non-textual content is essential.

To address these limitations, modern multimodal RAG systems have developed basic approaches to integrate and retrieve data across various modalities. The fundamental idea is to align different modalities into a shared embedding space for unified processing and retrieval. For example, models like CLIP\cite{radford2021learningtransferablevisualmodels} align vision and language in a shared space, while models for audio like Wav2Vec 2.0\cite{NEURIPS2020_92d1e1eb} and CLAP\cite{10095969} focus on audio-text alignment. For video, basic approaches like ViViT\cite{9710415} handle both spatial and temporal features. These foundational techniques provide the basis for more advanced multimodal RAG applications, which will be discussed in detail in the advanced approaches section.

\subsection{Knowledge Embedding}
Knowledge is stored in extensive textual documents, which are first segmented into concise and meaningful units, each encapsulating a distinct core idea. These units are then transformed into vector embeddings that encode semantic information, facilitating efficient retrieval through similarity metrics. In this subsection, we introduce chunk partitioning for segmenting knowledge into meaningful units and explore multimodal embedding models, including text, images, audio, and video. Table~\ref{tab:embedding} provides a summary of the corresponding models utilized in RAG.

\begin{table}[]
\caption{Representative knowledge embedding models categorized by knowledge form, encoder type, and embedding dimensionality.}
\label{tab:embedding}
\begin{tabular}{ccclc}
\hline
Knowledge Form                              & \multicolumn{3}{c}{Embedding Models}                                                                                                  & Dimensions                              \\ \hline
\multicolumn{1}{c|}{\multirow{20}{*}{Text}} & \multicolumn{1}{c|}{\multirow{3}{*}{Sparse Encoders}}                         & \multicolumn{2}{c|}{Bag of Words (BoW)}              & \multirow{3}{*}{Size of the Vocabulary} \\
\multicolumn{1}{c|}{}                       & \multicolumn{1}{c|}{}                                                         & \multicolumn{2}{c|}{N-gram}                          &                                         \\
\multicolumn{1}{c|}{}                       & \multicolumn{1}{c|}{}                                                         & \multicolumn{2}{c|}{TF-IDF}                          &                                         \\ \cline{2-5}
\multicolumn{1}{c|}{}                       & \multicolumn{1}{c|}{\multirow{3}{*}{Traditional Dense Encoders}}              & \multicolumn{2}{c|}{Word2vec\cite{NIPS2013_9aa42b31,DBLP:journals/corr/abs-1301-3781}}                        & Typically 100, 200, or 300              \\
\multicolumn{1}{c|}{}                       & \multicolumn{1}{c|}{}                                                         & \multicolumn{2}{c|}{GloVe\cite{pennington-etal-2014-glove}}                           & 50, 100, 200, or 300                    \\
\multicolumn{1}{c|}{}                       & \multicolumn{1}{c|}{}                                                         & \multicolumn{2}{c|}{fastText\cite{10.1162/tacl_a_00051}}                        & 300 (Default)                           \\ \cline{2-5}
\multicolumn{1}{c|}{}                       & \multicolumn{1}{c|}{\multirow{6}{*}{BERT Architecture Pre-training Encoders}} & \multicolumn{2}{c|}{BERT\cite{devlin-etal-2019-bert}}                            & 768 (Base)                              \\
\multicolumn{1}{c|}{}                       & \multicolumn{1}{c|}{}                                                         & \multicolumn{2}{c|}{RoBERTa\cite{DBLP:journals/corr/abs-1907-11692}}                         & 768 (Base)                              \\
\multicolumn{1}{c|}{}                       & \multicolumn{1}{c|}{}                                                         & \multicolumn{2}{c|}{ALBERT\cite{Lan2020ALBERT}}                          & 768 (Base)                              \\
\multicolumn{1}{c|}{}                       & \multicolumn{1}{c|}{}                                                         & \multicolumn{2}{c|}{DistilBERT\cite{DBLP:journals/corr/abs-1910-01108}}                      & 768                                     \\
\multicolumn{1}{c|}{}                       & \multicolumn{1}{c|}{}                                                         & \multicolumn{2}{c|}{ELECTRA\cite{Clark2020ELECTRA}}                         & 768 (Base)                              \\
\multicolumn{1}{c|}{}                       & \multicolumn{1}{c|}{}                                                         & \multicolumn{2}{c|}{DPR\cite{karpukhin-etal-2020-dense}}                             & 768                                     \\ \cline{2-5}
\multicolumn{1}{c|}{}                       & \multicolumn{1}{c|}{\multirow{4}{*}{LLM-based Encoders}}                      & \multicolumn{2}{c|}{BGE\cite{li2024makingtextembeddersfewshot}}                             & 4096                                    \\
\multicolumn{1}{c|}{}                       & \multicolumn{1}{c|}{}                                                         & \multicolumn{2}{c|}{NV-Embed\cite{lee2024nv}}                        & 4096                                    \\
\multicolumn{1}{c|}{}                       & \multicolumn{1}{c|}{}                                                         & \multicolumn{2}{c|}{SFR-Embedding\cite{SFR-embedding-2}}                   & 4096                                    \\
\multicolumn{1}{c|}{}                       & \multicolumn{1}{c|}{}                                                         & \multicolumn{2}{c|}{text-embedding\cite{text-embedding}}                  & 768                                     \\ \cline{2-5}
\multicolumn{1}{c|}{}                       & \multicolumn{1}{c|}{\multirow{4}{*}{Specific Domain Encoders}}                & \multicolumn{2}{c|}{\multirow{2}{*}{ru-en-RoSBERTa\cite{snegirev2024russianfocusedembeddersexplorationrumteb}}} & \multirow{2}{*}{1024}                   \\
\multicolumn{1}{c|}{}                       & \multicolumn{1}{c|}{}                                                         & \multicolumn{2}{c|}{}                                &                                         \\
\multicolumn{1}{c|}{}                       & \multicolumn{1}{c|}{}                                                         & \multicolumn{2}{c|}{USTORY\cite{10.1145/3539618.3591782}}                          & 768                                     \\
\multicolumn{1}{c|}{}                       & \multicolumn{1}{c|}{}                                                         & \multicolumn{2}{c|}{E5-Base-4k\cite{DBLP:journals/corr/abs-2404-12096}}                      & 768                                     \\ \hline
\multicolumn{1}{c|}{\multirow{2}{*}{Image}} & \multicolumn{3}{c|}{Vision Transformer (ViT)\cite{dosovitskiy2021an}}                                                                                        & 768 (Base)                              \\
\multicolumn{1}{c|}{}                       & \multicolumn{3}{c|}{CLIP\cite{pmlr-v139-radford21a}}                                                                   & 512                                     \\ \hline
\multicolumn{1}{c|}{\multirow{2}{*}{Audio}} & \multicolumn{3}{c|}{Wav2Vec 2.0\cite{NEURIPS2020_92d1e1eb}}                                                                                                     & 768 (Base)                              \\
\multicolumn{1}{c|}{}                       & \multicolumn{3}{c|}{CLAP\cite{10095969}}                                                                   & 512                                     \\ \hline
\multicolumn{1}{c|}{Video} & \multicolumn{3}{c|}{ViViT\cite{9710415}}                                                               & 768 (Base)             \\ \hline
\end{tabular}
\end{table}

\subsubsection{Chunk Partitioning }
Chunk partitioning is a fundamental process that directly influences retrieval quality and overall RAG performance \cite{shi2023largelanguagemodelseasily,yu2024chainofnoteenhancingrobustnessretrievalaugmented}. Its goal is to divide large texts into smaller, coherent “chunks” for efficient retrieval and context management \cite{wadhwa2024ragsrichparametersprobing}. Effective chunking must preserve semantic coherence while minimizing redundancy and information loss. Traditional methods, such as fixed-length, rule-based, or semantic-based partitioning, are simple but often fail to capture nuanced structures or cross-paragraph dependencies, leading to fragmented context and reduced retrieval accuracy \cite{lewis2021retrievalaugmentedgenerationknowledgeintensivenlp,zhang2021sequencemodelselfadaptivesliding,lyu2024crudragcomprehensivechinesebenchmark,xiao2024cpackpackedresourcesgeneral,Langchain,llamaindex}.

Recent advances move beyond fixed-length chunking toward adaptive, semantically informed strategies. Proposition-level chunking \cite{chen2024densexretrievalretrieval} segments text into atomic factual units, while LLM-driven approaches like LumberChunker \cite{duarte2024lumberchunkerlongformnarrativedocument} dynamically identify content shifts for context-sensitive partitioning. Meta-chunking methods \cite{zhao2024metachunkinglearningefficienttext}, including Margin Sampling and Perplexity Chunking, refine boundaries to balance granularity and coherence. Late chunking \cite{günther2024latechunkingcontextualchunk} embeds entire documents before segmentation, preserving global context, and LeCut \cite{10.1145/3477495.3531998} leverages attention mechanisms to determine optimal cut points based on retrieval relevance. Collectively, these adaptive techniques enable RAG systems to handle complex document structures more effectively, improving contextual fidelity and retrieval precision.

\subsubsection{Text Embedding Models}

During retrieval, RAG systems rely on vector similarity measures, such as cosine similarity, between queries and text chunks, making accurate vector representation essential for capturing semantic meaning. Early approaches such as Bag of Words, N-gram, and TF-IDF considered word frequency and limited context but struggled with sparsity and high dimensionality. Modern embeddings, like word2vec \cite{NIPS2013_9aa42b31,DBLP:journals/corr/abs-1301-3781}, GloVe \cite{pennington-etal-2014-glove}, and fastText \cite{10.1162/tacl_a_00051}, introduced distributed representations capturing contextual relationships, yet remained static. The transformer architecture \cite{NIPS2017_3f5ee243}, exemplified by BERT \cite{devlin-etal-2019-bert} and its successors RoBERTa \cite{DBLP:journals/corr/abs-1907-11692}, ALBERT \cite{Lan2020ALBERT}, and DPR \cite{karpukhin-etal-2020-dense}, revolutionized contextual embeddings and laid the foundation for RAG’s dense retrieval \cite{NEURIPS2020_6b493230}.

LLM-based embedders now dominate, leveraging large-scale pretraining for deeper semantic and contextual understanding. Models such as BGE \cite{li2024makingtextembeddersfewshot}, NV-Embed \cite{lee2024nv}, and SFR-Embedding \cite{SFR-embedding-2} achieve leading results across multilingual and domain-specific benchmarks. Domain-adapted variants like ru-en-RoSBERTa \cite{snegirev2024russianfocusedembeddersexplorationrumteb}, WE-iMKVec \cite{KHINE2024102758}, USTORY \cite{10.1145/3539618.3591782}, and E5-Base-4k \cite{DBLP:journals/corr/abs-2404-12096} extend these gains to specialized contexts, while LLM-Embedder \cite{zhang-etal-2024-multi-task} unifies general and task-specific retrieval through multi-task optimization and LLM-guided supervision.


Selecting appropriate embedding models and storage strategies remains crucial. Benchmarks such as MTEB \cite{muennighoff-etal-2023-mteb}, C-MTEB \cite{10.1145/3626772.3657878}, LongEmbed \cite{DBLP:journals/corr/abs-2404-12096}, SEB \cite{DBLP:journals/corr/abs-2406-02396}, and AIRBench \cite{air-bench} evaluate models across languages, contexts, and domains, with AIRBench specializing in RAG-oriented retrieval and re-ranking. In practice, vector databases like FAISS, Milvus, and Pinecone enable efficient similarity search and large-scale storage, serving as the backbone for scalable, high-precision retrieval in modern RAG systems.

\subsubsection{Multimodal Embedding Models}
Knowledge is represented not only in text but also in images, audio, and video. Consequently, there is an increasing demand for multimodal embedding models that integrate information from various modalities into a cohesive vector space. These models are specifically designed to capture the relationships and shared information across different data types, thereby enabling more comprehensive and unified representations.

For images, models process image formats such as JPG or PNG to generate embeddings within the same semantic vector space as text. Normalizer-Free ResNets (NFNet)\cite{pmlr-v139-brock21a} provides an efficient framework for extracting image features, while the Vision Transformer (ViT)\cite{dosovitskiy2021an} leverages the transformer architecture to learn high-quality representations. Contrastive Language-Image Pretraining (CLIP)\cite{pmlr-v139-radford21a} further advances the field by aligning visual and textual modalities through contrastive learning, producing versatile embeddings for applications like zero-shot classification and cross-modal retrieval. CLIP\cite{pmlr-v139-radford21a} has become a cornerstone of modern multimodal AI systems.

For audio, models extract essential features such as pitch, timbre, rhythm, and semantics, enabling effective and meaningful audio analysis for retrieval tasks. Wav2Vec 2.0\cite{NEURIPS2020_92d1e1eb}, a self-supervised learning model, learns audio representations directly from raw waveforms, producing rich, high-level embeddings suitable for diverse audio tasks. Inspired by CLIP\cite{pmlr-v139-radford21a}, Contrastive Language-Audio Pretraining (CLAP)\cite{10095969} is a SOTA model that generates audio embeddings by learning from paired audio and text data, offering a unified framework for integrating audio with natural language.

For video, models aim to represent video data in compact, feature-rich vectors that capture spatial, temporal, and semantic information. Video Vision Transformer (ViViT)\cite{9710415}, based on ViT\cite{pmlr-v139-brock21a}, can effectively handle video understanding tasks by capturing both spatial and temporal features. VideoPrism\cite{zhao2024videoprism} has also attracted attention for achieving state-of-the-art performance across a wide range of video understanding benchmarks. It is particularly notable for its ability to generalize well across different video domains without requiring task-specific fine-tuning.

In summary, high-quality embeddings are essential to the effectiveness and efficiency of a retriever. Recent text embedding models, such as BGE\cite{li2024makingtextembeddersfewshot} and NV-Embed\cite{lee2024nv}, perform strongly across a wide range of tasks. However, the rapid development of multimodal RAG continues to create demand for better multimodal embedding models.

\subsection{Knowledge Indexing}
In RAG, an index is defined as a structured organization of data that enables efficient access and retrieval of information from large-scale datasets. The index maps user queries to relevant document chunks, knowledge snippets, or other informational content, acting as a bridge between the stored data and the retrieval mechanism. The effectiveness of indexing is crucial for RAG systems, as it impacts response accuracy, retrieval speed, and computational efficiency.

\subsubsection{Index Structure}

\textit{Structured Index.} Structured indexing organizes data based on predefined, fixed attributes, typically following a tabular or relational format. In early knowledge retrieval efforts, such as in systems like \textbf{REALM}, textual inverted indexing was widely employed as a fundamental technique\cite{pmlr-v119-guu20a}, while Table RAG utilizes a table-specific indexing structure that combines column-based and row-based indices to efficiently retrieve relevant table entries for language generation tasks\cite{chen2024tablerag}.

\textit{Unstructured Index.} Unstructured indexing, in contrast, is designed for free-form or semi-structured data, which is more commonly used in modern RAG systems. Vector indexing uses vectors obtained from previous embedding stages to improve retrieval efficiency, such as naive RAG\cite{lewis2020retrieval}, ANCE\cite{xiong2020approximate} and G-retriever which uses language models to convert textual attributes of graphs into vectors\cite{he2024g}.


Graph indexing, an unstructured indexing approach, exploits graph structures to represent and retrieve interconnected data. In this paradigm, nodes correspond to data points and edges encode their relationships, allowing the index to naturally capture semantic and contextual dependencies, an essential capability for RAG retrieval. Owing to these strengths, graph-based indexing is increasingly applied in RAG systems. For example, ToG employs knowledge graphs as its core index \cite{sun2023think}, with ToG 2.0 extending this design to incorporate unstructured text \cite{ma2024think}. Graph RAG \cite{edge2024local} extracts entities and relations via LLMs, partitions the graph into sub-communities, and generates summaries for each. RAPTOR \cite{sarthi2024raptor} builds hierarchical trees of recursive embeddings and summaries to improve long-text comprehension, while GRAFT-Net and PullNet \cite{sun2018open,sun2019pullnet} integrate textual and graph data into heterogeneous networks for reasoning and subgraph extraction. Once constructed, these graphs support efficient retrieval through algorithms such as BFS and HNSW \cite{malkov2018efficient}, enabling scalable and semantically rich knowledge access within RAG frameworks.

\textit{Hybrid Index.}
In practice, a hybrid indexing approach generally performs more effectively than a single indexing method, as it integrates both structured and unstructured techniques to optimize retrieval across a wide variety of data types \cite{peng2024graph}. ColBERT employs an inverted index structure to swiftly filter through relevant documents, after which the similarity between the query and the candidate document embedding is computed using a vector index structure. This dual approach ensures high semantic accuracy while also facilitating rapid retrieval \cite{khattab2020colbert}.

\subsubsection{Index Updating \& Storage}
Although less frequently discussed, index updating plays a crucial role in ensuring that RAG systems remain both accurate and efficient when new data is introduced. In practice, index updates are typically carried out through two primary approaches: incremental updates and periodic reconstruction. Among these, incremental updates, the process of adding or modifying entries within an existing index without the need to rebuild the entire structure, allow for efficient adaptation to evolving data while minimizing computational overhead. This makes incremental updates particularly advantageous in dynamic systems, where new information is constantly introduced, such as Light RAG \cite{guo2024lightrag} and REALM \cite{pmlr-v119-guu20a}.

Moreover, it is the storage of indexes that also plays a fundamental role in RAG models, which is usually influenced by the structure of indexes and will, directly or indirectly, influence the performance of the retrieval step. For vector indexes, Chroma utilizes a distributed architecture and is horizontally scalable for mass storage, Faiss has excellent similarity search capabilities, providing an optimized indexing method for efficient retrieval of nearest neighbors\cite{johnson2019billion} and Weaviate uses graph structures for its data storage and integrates machine learning models to optimize vector storage and search\cite{weaviate2024}.For graph indexes, there are many well-established graph databases available, such as JanusGraph and HugeGraph, which use Gremlin as their query language\cite{janusgraph2024, hugegraph2024}, as well as Neo4j, the largest and most recognizable one which uses Cypher\cite{neo4j2024}.

\subsection{Knowledge Retrieval}
Knowledge Retrieval is the process of identifying and retrieving relevant knowledge from a vector database based on a given query. This process includes both retrieval strategies, which focus on how to identify relevant knowledge, and search methods, which define the algorithms used to quickly locate the most relevant information.

\begin{table}[ht]
    \caption{Representative knowledge retrieval models categorized by retrieval strategy, search approach, and deep learning usage.}
    \label{tab:retrieval}
    \centering
    \begin{tabular}{c|c|c|c}
    \hline
         &  Type & Models & Deep Learning\\
    \hline
         \multirow{13}{*}{Retrieval Strategies} & \multirow{6}{*}{Sparse Retrieval} & BM25\cite{robertson2009probabilistic} & $\times$ \\
         &  & TF-IDF\cite{robertson1997relevance} & $\times$ \\
         &  & likelihood \cite{lafferty2001document} & $\times$ \\
         & & uniCOIL  \cite{lin2021few} & \checkmark\\
         & & TILDE\cite{zhuang2021tilde} & \checkmark \\
         & & SPLADE\cite{formal2021splade} & \checkmark\\
         \cline{2-4}
         & \multirow{3}{*}{Dense Retrieval} & DPR \cite{karpukhin2020dense} & \checkmark \\
         & & ANCE \cite{xiong2020approximate} & \checkmark \\
         & & Llama2vec \cite{li2024llama2vec} & \checkmark \\
         & & RepLLaMA \cite{ma2024fine} & \checkmark \\
         \cline{2-4}
         & \multirow{4}{*}{Hybrid Retrieval} & RAP-Gen\cite{wang2023rap} & \checkmark \\
         & & BlendedRAG\cite{sawarkar2024blended} & \checkmark\\
         & & ReACC\cite{lu2022reacc} & \checkmark\\
         & & BASHEXPLAINER\cite{yu2022bashexplainer} & \checkmark\\
    \hline
        \multirow{10}{*}{Search Approaches} & \multirow{3}{*}{Nearest Neighbor Search} & Brute Force Algorithm & $\times$\\
        & & KD Tree \cite{bentley1975multidimensional} & $\times$\\
        & & Ball Tree \cite{dolatshah2015ball} & $\times$\\
        & & M-tree\cite{ciaccia1997m} & $\times$\\
        \cline{2-4}
        & \multirow{7}{*}{Approximate Nearest Neighbor Search} & Locality-Sensitive Hashing\cite{andoni2008near,datar2004locality} & $\times$\\
        & & Spectral Hashing\cite{weiss2008spectral} & $\times$\\
        & & Deep Hashing\cite{liu2016deep} & \checkmark\\
        & & K-means tree\cite{tavallali2021k} &$\times$\\
        & & ANNOY & -\\
        & & HNSW \cite{malkov2018efficient} & $\times$\\
        & & Product Quantization\cite{jegou2010product} & $\times$\\
    \hline
    \end{tabular}
\end{table}

\subsubsection{Retrieval Strategies}
The objective of retrieval is to identify and extract the most relevant knowledge based on the input query. Specifically, the task involves retrieving the top-k most relevant chunks by employing a similarity function. Based on different similarity functions, retrieval strategies can be categorized into into three types: sparse retrieval, dense retrieval, and hybrid retrieval.

\textit{Sparse Retrieval.} Sparse retrieval strategies leverage sparse vectors to retrieve documents or chunks through term analysis and matching. Traditional sparse retrieval strategies utilize term matching metrics such as BM25\cite{robertson2009probabilistic}, TF-IDF\cite{robertson1997relevance} and query likelihood\cite{lafferty2001document}, which estimate the relevance of documents to queries by calculating the frequency of term occurrences and inverse document frequency. The language modeling approach~\cite{ponte2017language,zhai2004study} further interprets retrieval as estimating the likelihood of a query being generated by a document-specific language model, bridging traditional IR and generative modeling. Owing to their interpretability, robustness, and computational efficiency, these methods remain widely adopted in RAG pipelines for initial candidate selection and hybrid retrieval setups. Recently, some research explored the use of transformer language models as sparse retrievers. Transformer-based sparse retrievers such as uniCOIL~\cite{lin2021few}, TILDE~\cite{zhuang2021tilde}, and SPLADE~\cite{formal2021splade} have extended bag-of-words representations with contextualized term weights, combining the scalability of classical sparse methods with the semantic awareness of neural encoders, and establishing sparse retrieval as a strong and efficient foundation for modern RAG systems.

\textit{Dense Retrieval.} Dense retrieval strategies encode both queries and documents into a low-dimensional vector space, where relevance is measured by the dot product or cosine similarity between their vector representations. Early neural retrieval architectures, such as DSSM~\cite{huang2013learning}, C-DSSM~\cite{shen2014learning}, and KNRM~\cite{xiong2017end}, first demonstrated the feasibility of neural text matching by leveraging feedforward, convolutional, or interaction-based mechanisms. These models inspired the development of dual-encoder architectures such as DPR~\cite{karpukhin2020dense} and ANCE~\cite{xiong2020approximate}, which employ transformer encoders and contrastive objectives to align queries and documents. Through large-scale pretraining and hard negative mining, dense retrievers achieved strong semantic generalization. Recent research extends dense retrieval into the era of large language models (LLMs). With their superior contextual reasoning and representational capacity, LLMs enable discriminative embedding generation and adaptive retrieval optimization~\cite{muennighoff2202sgpt,neelakantan2022text,zhang2023language,ma2024fine,li2024llama2vec}. For instance, Llama2Vec~\cite{li2024llama2vec} introduces unsupervised embedding alignment to adapt pretrained LLMs for retrieval, while RepLLaMA~\cite{ma2024fine} fine-tunes LLaMA for long-document encoding and efficient retrieval. Moreover, learned vector quantization and compression approaches such as RepCONC~\cite{10.1145/3488560.3498443} and JPQ~\cite{10.1145/3459637.3482358} further enhance scalability without compromising retrieval quality. Overall, dense retrieval has evolved from early neural architectures to LLM-based embeddings, forming a crucial component of modern RAG pipelines by bridging deep semantic understanding and scalable knowledge access.


\textit{Hybrid Retrieval.} Hybrid retrieval strategies combine both sparse and dense retrieval techniques, aiming to optimize performance by leveraging the strengths of each approach. RAP-Gen\cite{wang2023rap} and BlendedRAG\cite{sawarkar2024blended} integrate traditional keyword matching with deep semantic understanding, allowing systems to benefit from the efficiency of sparse retrieval while capturing deeper context through dense representations. BASHEXPLAINER\cite{yu2022bashexplainer} employs a two-stage training strategy that first utilizes a dense retriever to capture semantic information, followed by a sparse retriever to acquire lexical information, resulting in automatic code comment generation that performs well. This dual strategy addresses the limitations of each individual method; for instance, sparse strategies may struggle with semantic nuances, while dense strategies can be computationally intensive. By leveraging the strengths of both, hybrid models enhance retrieval accuracy and relevance across various tasks.

\subsubsection{Search Approaches}
Search approaches refer to algorithms designed to efficiently identify similar vectors from the vector database for a given query vector. Search approaches can be divided into two types: Nearest Neighbor Search (NNS) and Approximate Nearest Neighbor Search (ANNS).  NNS aims to find the vector in a given set that is closest (or most similar) to the query vector, while ANNS is a variation of NNS that allows for a controlled degree of error or approximation in the search results\cite{han2023comprehensive}.

\textit{Nearest Neighbor Search.} The brute force algorithm for NNS is a straightforward algorithm that exhaustively scans all vectors in the database, calculating the distance to the query vector to identify the closest one. However, this method is computationally expensive and impractical for large-scale datasets. To mitigate this limitation, tree-based methods have been introduced to enhance search efficiency. For example, Bentley\cite{bentley1975multidimensional} proposed a method based on the k-d tree, which recursively partitions k-dimensional space into hyperrectangular regions, improving both data organization and search speed. Other tree-based structures, such as Ball-tree\cite{omohundro1989five,dolatshah2015ball,liu2006new}, R-tree\cite{guttman1984r}, and M-tree\cite{ciaccia1997m} have also been used to enhance nearest neighbor search by partitioning the data into variou structures, such as hyperspheres, rectangles, or metric spaces, thereby improving search performance, particularly in high-dimensional and complex datasets.

\textit{Approximate Nearest Neighbor Search.} ANNS strike a balance between accuracy, speed, and memory efficiency, making it particularly useful for large-scale and high-dimensional data\cite{han2023comprehensive}.This involves hash-based methods, tree-based methods, graph-based methods, and quantization-based methods. Hash-based methods, such as Locality-Sensitive Hashing\cite{andoni2008near,datar2004locality}, Spectral Hashing\cite{weiss2008spectral}, and Deep Hashing\cite{liu2016deep}, convert high-dimensional vectors into binary codes, optimizing memory usage and accelerating search operations. For instance, Deep Hashing\cite{liu2016deep} employs deep neural networks to learn hash functions that map high-dimensional vectors into binary codes while preserving the semantic relationships between similar data. Tree based ANNS methods, including K-means tree\cite{tavallali2021k} and ANNOY), organize data hierarchically, reducing the search space by efficiently traversing tree structures. These methods divide the dataset into partitions or clusters, such that only relevant regions are explored during the search. Graph-based methods, such as Hierarchical Navigable Small World (HNSW) \cite{malkov2018efficient}, connect data points by edges that reflect their proximity to one another, allowing for fast nearest neighbor searches by navigating through the graph.  Finally, quantization-based methods, such as Product Quantization\cite{jegou2010product}, aim to compress the data by quantizing the vectors into a smaller codebook, thus reducing memory requirements while maintaining a good balance between search speed and accuracy. Overall, these diverse ANNS approaches provide powerful solutions for fast and efficient nearest neighbor search in large-scale, high-dimensional datasets, each with its own trade-offs in terms of accuracy, speed, and memory usage.

\subsection{Knowledge Integration}
Knowledge Integration is the process of fusing retrieved external knowledge with the generator's internal knowledge to improve output accuracy and coherence. The method chosen for this fusion represents a critical design decision in RAG architectures, with different strategies offering distinct trade-offs between model intrusiveness, computational complexity, and representational power. These strategies can be broadly categorized based on the architectural stage at which integration occurs: the input layer, intermediate layers, or the output layer.

\subsubsection*{Input Layer Integration}
Integrating knowledge at the input layer is the most direct and model-agnostic approach, enhancing the initial context provided to the generator. These techniques can be distinguished by whether they operate on raw text or on feature representations.

Text-level integration is the most basic method, typically involving the straightforward concatenation of the top-k retrieved documents with the original query. However, this simple approach can introduce noise and may exceed the model's context window. To address these limitations, subsequent research has concentrated on refining the concatenated context. Some methods focus on improving signal quality by reranking chunks to prioritize the most relevant content \cite{glass2022re2g, li2024enhancing}, or by applying weighted filtering techniques to remove irrelevant information \cite{lyu2023improving, arefeen2024leancontext}. Other methods tackle the input length constraint by compressing the context, thereby enabling the model to process more information within a fixed window size \cite{liu2023tcra, jiang2023longllmlingua, xu2023recomp, jin2024biderbridgingknowledgeinconsistency}. These compression techniques are designed to preserve essential information while reducing token count, facilitating more efficient generation.

In contrast to manipulating raw text, feature-level integration operates on the encoded representations of the query and retrieved documents. This method transforms both inputs into dense or sparse vectors before feeding them into the model \cite{izacard2020leveraging,de2023pre}. By working in the feature space, this approach offers more flexible and fine-grained control over how external knowledge is combined with the original input, moving beyond simple textual concatenation.

\subsubsection*{Intermediate Layer Integration}
A more integrated approach injects external knowledge directly into the generator’s hidden layers, allowing the model to condition its internal representations on retrieved information during generation. This method requires architectural modifications, resulting in more tightly coupled systems.

A dominant strategy within this category is the use of attention mechanisms. The RETRO model \cite{borgeaud2022improving}, for instance, exemplifies this by introducing a dedicated cross-attention module that merges retrieved information with the model's intermediate representations at each layer. Building on similar principles, other models introduce specialized memory structures. TOME \cite{de2021mention} employs a \textit{Mention Memory} to store and retrieve entity representations, while LongMem \cite{de2021mention} utilizes an adaptive residual network to efficiently access long-term memory. These attention-based methods provide a powerful, context-aware fusion mechanism but come at the cost of increased computational complexity.

As a more lightweight alternative to attention, other methods incorporate knowledge via simple weighted additions \cite{fevry2020entities}. These techniques learn a weight for the embeddings of the top-k retrieved documents and then sum these weighted embeddings into the model's intermediate layers. This offers a computationally cheaper solution for intermediate fusion, though it may lack the dynamic, token-level specificity of cross-attention.

\subsubsection*{Output Layer Integration}
The least intrusive integration strategy modifies the model's final output distribution by incorporating knowledge at the logit level. This post-processing approach allows the external knowledge to directly influence the final token probabilities without altering the core generator architecture.

One branch of this approach is ensemble-based integration, which combines the model's output distribution with one derived from the retrieved neighbors. The canonical example, kNN-LM \cite{khandelwal2019generalization}, interpolates the probabilities of the k-nearest neighbors from an external datastore with the language model's own predictions. This effectively broadens the model's knowledge at the point of decision, improving generalization and robustness.

A second branch, calibration-based integration, uses retrieved knowledge not just to supplement but to refine the model’s prediction confidence. Rather than a simple interpolation, these methods use retrieval-based logits to adjust or calibrate the model's output. This technique is evident in confidence-enhanced kNN-MT \cite{jiang2022towards} for machine translation, EDITSUM \cite{li2021editsum} for code summarization, and MA \cite{fei2021memory} for image captioning. These methods use external information as a form of verification or correction for the generator's final prediction.

\subsubsection*{Contrastive Decoding as Post-hoc Integration}

Beyond ensemble- and calibration-based strategies, recent research has explored a novel inference-time approach known as \textit{Contrastive Decoding} (CD), which provides a principled mechanism for balancing the language model’s internal (parametric) knowledge with external (retrieved) knowledge without additional training \cite{shi-etal-2024-trusting,zhao-etal-2024-enhancing}.
Unlike conventional output-layer fusion that linearly interpolates probability distributions, CD explicitly contrasts the model’s token-level likelihoods under two different conditions: with and without retrieved context. Formally, the adjusted output probability at decoding step $t$ can be expressed as:
\[
\tilde{p}(y_t|c,x,y_{<t}) \propto p(y_t|c,x,y_{<t}) \left( \frac{p(y_t|c,x,y_{<t})}{p(y_t|x,y_{<t})} \right)^{\alpha},
\]
where $\alpha$ controls the degree of contrastive adjustment. This formulation effectively penalizes generation paths that deviate from external evidence, helping to mitigate hallucinations and factual errors in knowledge conflict scenarios.

Building upon this foundation, \cite{zhao-etal-2024-enhancing} introduce a multi-input extension that integrates both relevant and irrelevant retrievals, encouraging the model to discriminate between supportive and misleading evidence. Such contrastive formulations can be viewed as a post-hoc, logit-level knowledge alignment mechanism that dynamically calibrates the generator’s output distribution based on contextual reliability.

From the perspective of knowledge integration, Contrastive Decoding complements the seamless knowledge fusion objective discussed above. Both aim to reconcile discrepancies between the model’s internal and external knowledge sources; however, while most integration strategies modify model representations or architectures, CD operates purely at inference time. Consequently, it offers a lightweight yet effective means of achieving factual consistency and conflict resolution in RAG without architectural modification or retraining.

\subsection{Answer Generation}
The generation component in RAG systems serves as the final synthesis engine, tasked with producing responses that are both accurate and contextually relevant. Its performance hinges on two core capabilities: filtering irrelevant or contradictory retrieved data (Denoising) and synthesizing information to form a coherent whole (Reasoning). The choice and implementation of these capabilities are highly dependent on the target application, representing a trade-off between fidelity to retrieved knowledge and the model's intrinsic generative abilities. This section examines how different strategies for Denoising and Reasoning adapt to various models and tasks to improve generation quality.

\begin{figure*}
    \centering
    \includegraphics[width=1\linewidth]{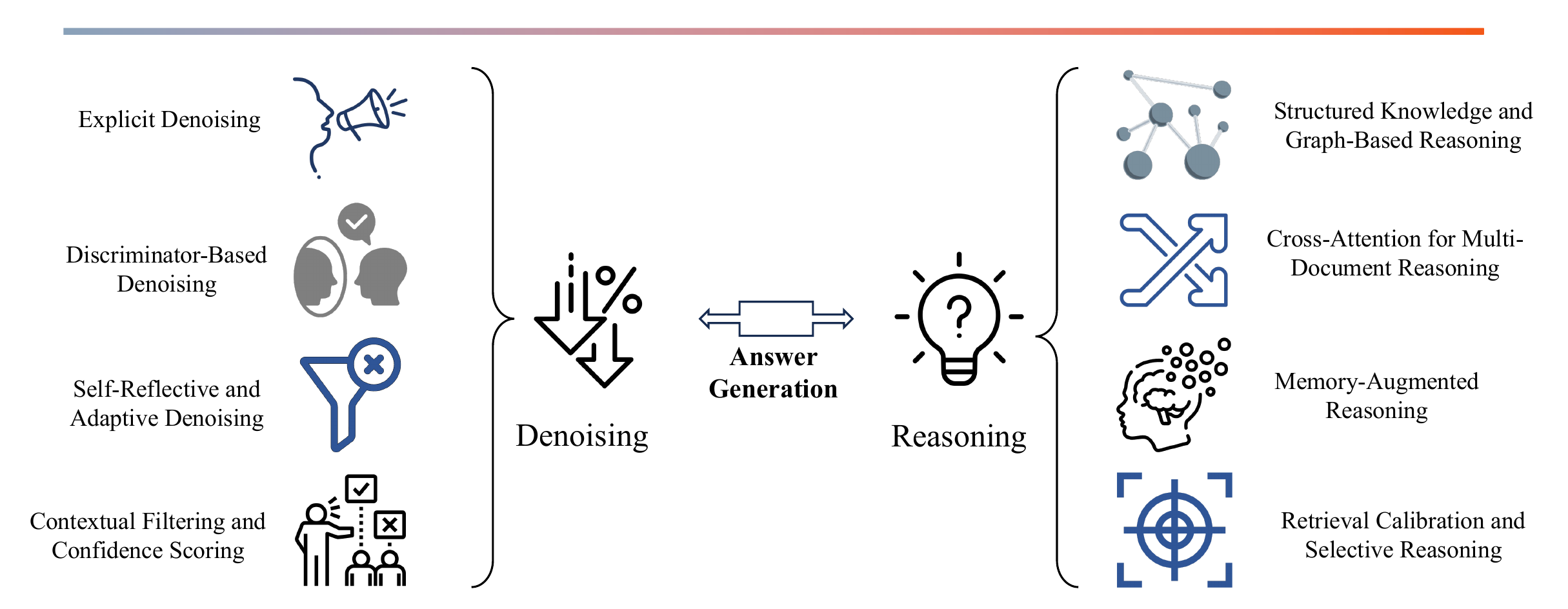}
    \caption{Answer generation strategies utilized by RAG, including Denoising and Reasoning.}
    \label{fig:answer_generation}
\end{figure*}

\subsubsection{Denoising}
Denoising is a critical pre-condition for high-quality generation, as it mitigates the influence of irrelevant or contradictory information. The specific denoising strategy often depends on the task's tolerance for noise and whether the core language model can be modified. These techniques can be categorized by their point of intervention: explicit, model-driven filtering; the use of external discriminator models; iterative self-correction; or simple confidence-based pre-filtering.

\textit{Explicit Denoising Techniques.}
One powerful, albeit computationally intensive, approach is to leverage the LLM itself for denoising through explicit supervision. For tasks requiring high interpretability, InstructRAG \cite{wei2024instructrag} instructs the model to generate intermediate rationales, which act as a filter by clarifying the relevance of each document. In a similar vein, the REFEED framework \cite{yu2023improving} is well-suited for fact-checking applications, as it uses an iterative loop where the LLM refines answers by reassessing factual accuracy. For conversational tasks, where context can be noisy, methods like COTED \cite{10.1145/3477495.3531961} apply contrastive learning to denoise irrelevant dialogue turns, proving effective in low-data scenarios.

\textit{Discriminator-Based Denoising.}
As an alternative to using the generator for filtering, a discriminator-based approach outsources this task to a separate, specialized model. This is particularly useful in production systems where a standardized quality gate is needed before the final generation phase. The COMBO framework \cite{zhang2023merging}, for instance, employs a pretrained discriminator to assess the coherence of generated-retrieved paragraph pairs, ensuring that contradictory information is filtered to minimize hallucinations.

\textit{Self-Reflective and Adaptive Denoising.}
A more dynamic and adaptive strategy involves empowering the model to critique and revise its own output. Self-RAG \cite{asai2023self} exemplifies this by introducing a self-reflection mechanism, allowing the model to handle queries of varying complexity where the relevance of information is not immediately obvious. This introspective capability is further extended by MetaRAG \cite{10.1145/3589334.3645481}, which integrates metacognition to help the model plan and evaluate its response strategies. These methods are best suited for complex, open-ended tasks where a fixed filtering policy is insufficient. Adaptive retrieval strategies \cite{jeong2024adaptive} complement this by allowing the model to dynamically adjust its retrieval scope based on the task, as illustrated in Figure \ref{fig:answer_generation}.

\textit{Contextual Filtering and Confidence Scoring.}
The most lightweight, model-agnostic approach involves pre-filtering based on confidence scores. Models assign scores to retrieved documents according to their alignment with the query, and low-confidence documents are discarded before integration \cite{njeh2024enhancing}. This approach is highly efficient and valuable for open-domain question answering, where retrieval quality is variable and a fast, scalable filtering method is essential.

These denoising methods represent a spectrum of strategies, from lightweight pre-filtering suitable for real-time applications to computationally intensive self-correction loops designed for high-stakes, complex reasoning tasks.

\subsubsection{Reasoning}
Once high-quality information is isolated, the generator must synthesize it. Effective reasoning enables the model to connect disparate facts, understand complex relationships, and produce answers that are more than the sum of their parts. The appropriate reasoning strategy is determined by the nature of the task, whether it requires navigating structured relationships, synthesizing long-form text, or performing multi-step logical inference.

\textit{Structured Knowledge and Graph-Based Reasoning.}
For domains where entities and their relationships are well-defined, such as scientific literature or enterprise knowledge bases, integrating structured knowledge provides a powerful reasoning backbone. Think-on-Graph 2.0 \cite{ma2024think} leverages knowledge graphs alongside unstructured text, enabling the model to perform complex relational reasoning that would be difficult to infer from text alone. This is ideal for expert systems or any task requiring deep domain-specific knowledge.

\textit{Cross-Attention for Multi-Document Reasoning.}
Unlike approaches relying on pre-existing structure, cross-attention mechanisms enable models to discover connections dynamically within unstructured text. This architectural design is particularly effective for open-domain synthesis tasks, such as generating summaries from multiple articles. The RETRO model \cite{borgeaud2022improving} employs chunked cross-attention to attend to relevant information across retrieved chunks, while kNN-based methods \cite{khandelwal2019generalization} use attention to integrate neighborhood information, enabling fluid reasoning across documents.

\textit{Memory-Augmented Reasoning.}
For tasks requiring longitudinal consistency, such as conversational AI or biographical summarization, memory-augmented models are essential. Models like EAE \cite{fevry2020entities} and TOME \cite{de2021mention} integrate entity-specific memory modules that store and retrieve information, allowing the model to maintain a coherent understanding of an entity over a long context.

\textit{Retrieval Calibration and Selective Reasoning.}
A more advanced form of reasoning involves the model learning *which* retrieved information is most critical to the task at hand. Retrieval calibration techniques \cite{jiang2021can} train the model to prioritize certain facts, which is vital for focused question answering where a single key piece of information may be buried among less relevant details. Selective reasoning frameworks \cite{baek2024probing} take this further by enabling the model to evaluate information in stages, mimicking a more human-like process of deliberation.

\textit{Hierarchical and Multi-Pass Reasoning.}
Complex queries often require multi-step inference that cannot be resolved in a single pass. Hierarchical or multi-pass reasoning models \cite{qiu2024learning} address this by processing information iteratively. This strategy is indispensable for tasks involving causal chains, such as questions about the economic consequences of an event, or comparative analysis, allowing the model to build a layered understanding by revisiting evidence with new information. In RAG systems, such reasoning can decompose an initial query into sub-questions, retrieve evidence for each intermediate step, and refine subsequent retrieval based on partial conclusions. 

Overall, the choice of denoising and reasoning techniques dictates a RAG model's capabilities and its suitability for different applications. Denoising ensures informational fidelity, while reasoning provides analytical depth. Future advancements will likely focus on more tightly integrating these processes, creating RAG systems that are not only accurate but also capable of autonomous and robust reasoning in complex, real-world scenarios.

\subsection{Knowledge Citation}


Citation in RAG is crucial for ensuring transparency, trustworthiness, and factual grounding of the model's responses. By attributing generated content to verifiable sources, it allows users to easily validate the information,
reduces the burden of claim verification, and improves the evaluation process. Additionally, effective citation helps mitigate hallucinations, reinforcing the factual integrity of the model's outputs.  In this subsection, we introduce strategies for generating citations in RAG and explore the evolving granularity of citations.
\subsubsection{Strategies for Citation Generation}


There are two main strategies for generating citations in language models: simultaneous citation generation and post-generation citation retrieval. Simultaneous generation, used by models like WebGPT\cite{nakano2022webgptbrowserassistedquestionansweringhuman}, GopherCite\cite{menick2022teachinglanguagemodelssupport}, and RECLAIM\cite{xia2024groundsentenceimprovingretrievalaugmented}, retrieves information in real time during response generation. This method ensures that the answer and citation are closely aligned, reducing hallucinations and improving factual accuracy. By contrast, post-generation citation, used by models like RARR\cite{gao2023rarrresearchingrevisinglanguage} and LaMDA\cite{thoppilan2022lamdalanguagemodelsdialog}, involves generating the answer first and retrieving citations afterward. While this approach reduces computational complexity, it increases the risk of discrepancies between the response and the cited sources, as the answer is produced independently of the citations. Both approaches offer unique benefits: simultaneous generation provides stronger factual grounding, while post-generation citation offers greater flexibility in response generation.

\subsubsection{Advances in Citation Granularity}
Citation granularity, the level of detail in a citation, has improved markedly in recent models. Early systems like LaMDA used coarse-grained references, often citing entire documents or URLs, which, while helpful for factual grounding, forced users to sift through irrelevant information. Models such as WebGPT, WebBrain \cite{qian2023webbrainlearninggeneratefactually}, and GopherCite have advanced toward fine-grained citations, retrieving specific snippets of evidence or interpreting long documents to support individual claims. RECLAIM exemplifies the highest granularity, linking each claim to exact sentences in the source. Similarly, LongCite \cite{zhang2024longciteenablingllmsgenerate} introduced CoF (Coarse to Fine), a pipeline using existing LLMs to generate precise sentence-level citations. Think\&Cite formulates attribution-based sentence generation as a multi-hop reasoning problem, solved through Self-Guided Monte Carlo tree search \cite{li2024think}. This progression improves fact-checking efficiency by reducing user effort and ensuring information is easily traceable to its source.

\section{Advanced RAG Approaches}
\label{sec:advanced}
Advanced RAG approaches, defined as methods beyond basic RAG models, encompass a range of cutting-edge techniques designed to overcome the limitations of basic RAG systems. These approaches aim to enhance RAG systems across multiple dimensions, including training optimization, multimodal processing, memory enhancement, agentic reasoning, and parameter-level knowledge integration. Specifically, this section delves into five key advancements: RAG Training, which focuses on improving the synergy between retrieval and generation; Multimodal RAG, which integrates multiple sensory modalities for enriched outputs; Memory RAG, which incorporates long-term memory to improve contextual reasoning and personalization; Agentic RAG, which introduces iterative and dynamic optimization mechanisms to handle evolving information needs; and Parametric RAG, which integrates retrieved knowledge into model parameters to reduce online computation and improve reasoning efficiency. Together, these approaches expand RAG’s capabilities, enabling it to address complex, dynamic, and specialized tasks.

\subsection{RAG Training}

\begin{figure*}
    \centering
    \includegraphics[width=.9\linewidth]{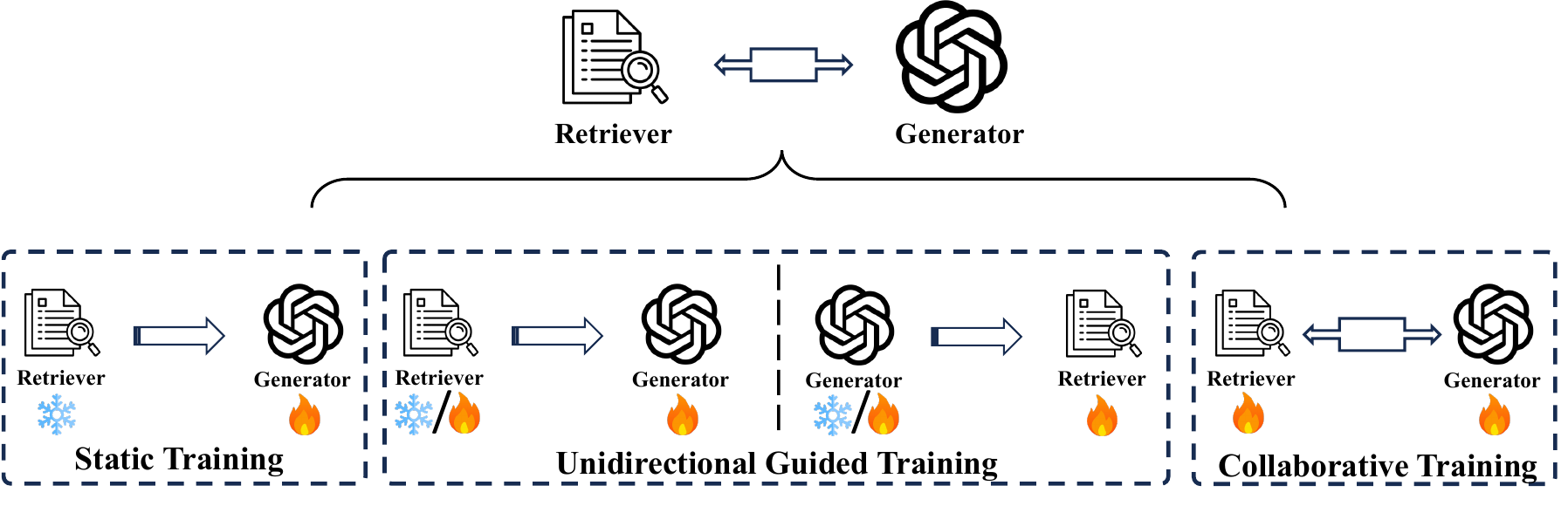}
    \caption{RAG training includes static training, unidirectional guided training, and collaborative training.}
    \label{fig:rag_training}
\end{figure*}

Training RAG models requires balancing the optimization of both retrieval and generation components to achieve optimal performance. Effective training strategies ensure the retriever fetches relevant information while the generator produces coherent and accurate outputs. This section reviews various methods in RAG training, including static training, unidirectional guided training, and collaborative training, shown in Figure \ref{fig:rag_training}. Each approach offers distinct benefits and challenges, affecting the effectiveness and adaptability of RAG models across applications. By exploring these paradigms, we can enhance the integration of retrieval and generation processes, ultimately improving RAG performance.

\subsubsection{Static Training}
Static optimization training is a simple yet effective approach where either the retriever or generator remains fixed during training, allowing optimization to focus on the other. This method is useful in limited resources or when rapid deployment is required. For instance, fixing the retriever while optimizing the generator lets the system benefit from established retrieval methods such as BM25 \cite{robertson2009probabilistic} or pre-trained models like BERT \cite{devlin2018bert}, without the overhead of joint training. This yields faster training cycles and lower resource use. However, its main drawback is the potential loss of synergy. Since only one component is optimized, interaction between retrieval and generation may not be fully realized, limiting adaptability to specific tasks or domains. Mitigating this requires careful selection of the fixed component and optimization strategy to ensure the evolving module effectively leverages fixed knowledge.

\subsubsection{Unidirectional Guided Training}
Unidirectional guided training introduces a directional optimization strategy where one component directs the training of the other. This approach can be categorized into two distinct methodologies: retriever-guided generator training and generator-guided retriever training.

\textit{Retriever-Guided Generator Training.}
In retriever-guided generator training, the retriever plays a central role in shaping generator optimization. By leveraging high-quality retrieved documents, the generator is fine-tuned to better integrate and utilize external information. For example, RETRO~\cite{borgeaud2022improving} employs a pre-trained BERT retriever to provide relevant context that enhances generation quality. Similarly, RALMs~\cite{yoran2023making} use a pre-trained COLBERTv2 retriever to fine-tune large language models (LLMs), improving their ability to incorporate retrieved data effectively. ITER-RTGEN~\cite{shao2023enhancing} leverages S-BERT to guide T5 fine-tuning, ensuring generated text aligns closely with retrieved information. Moreover, SMALLCAP~\cite{ramos2023smallcap} integrates CLIP as a retriever to guide GPT-2 in producing accurate, contextually appropriate captions. This guided approach ensures the generator consistently benefits from relevant information, enhancing the overall quality and relevance of generated content.

\textit{Generator-Guided Retriever Training.}
Conversely, generator-guided retriever training focuses on optimizing the retriever based on the generator's performance and requirements. In this paradigm, the generator's ability to produce coherent and accurate text influences the retriever's selection process. DKRR \cite{izacard2020distilling} leverages the generator's attention scores to fine-tune the retriever, enhancing its capability to select the most pertinent information. AAR \cite{yu2023augmentation} employs smaller language models to generate supervision signals that guide the retriever's training, ensuring that the retrieved documents are optimally aligned with the generator's needs. RA-DIT \cite{lin2023ra} fine-tunes large language models before training the retriever, fostering better alignment and synergy between the two components. Additionally, UPRISE \cite{cheng2023uprise} uses a frozen LLM to guide the fine-tuning of a prompt retriever, thereby improving its effectiveness in retrieving data that the generator can utilize more effectively. This bidirectional influence ensures that the retriever evolves in tandem with the generator, fostering a more integrated and efficient RAG system.

\subsubsection{Collaborative Training}
Collaborative optimization training adopts a co-optimization strategy, simultaneously refining both the retriever and the generator to achieve optimal overall system performance. This integrated approach ensures that improvements in one component enhance the other, fostering a relationship that maximizes the capabilities of the entire RAG system. By training both components, the system can better align the retrieval of relevant documents with the generation of accurate and coherent text. Notable implementations of this approach include RAG \cite{lewis2020retrieval} and the work by \cite{guu2020retrieval}, both of which utilize a joint training paradigm combined with Maximum Inner Product Search (MIPS) \cite{ram2012maximum, ding2019bilinear} to effectively optimize the retrieval process. This integrated training methodology enables the retriever to refine its relevance scoring based on the generator's feedback, while the generator simultaneously learns to leverage the retrieved information more effectively. The outcome is a more robust and adaptive RAG system, capable of handling a broader range of tasks and adapting to diverse input contexts with greater precision and fluency.

\subsection{Multimodal RAG}

Multimodal RAG extends traditional text-based RAG systems by incorporating multiple modalities including image, audio, video, etc. Compared to text-based RAG systems, multimodal RAG faces two main challenges. First, it needs to effectively represent and retrieve knowledge across different modalities, requiring sophisticated approaches to transform diverse data types into unified representations and enable cross-modal search. Second, after retrieving multimodal knowledge, the system must understand and utilize information across modalities to generate appropriate responses, demanding advanced techniques for multimodal understanding and generation. This section explores these two aspects: how multimodal RAG systems represent and retrieve diverse types of information, and how they understand and integrate this multimodal knowledge to produce contextually grounded responses.

\subsubsection{Multimodal Representation and Retrieval}

The foundation of multimodal RAG systems lies in their ability to effectively represent and retrieve information across different modalities. This requires both sophisticated embedding models to transform diverse data types into unified vector representations and specialized retrieval mechanisms to efficiently search across modalities. Various approaches have been developed to address these fundamental challenges:

Different modalities require specialized approaches for effective representation and retrieval. For visual content, models like NFNet\cite{pmlr-v139-brock21a} and Vision Transformer (ViT)\cite{dosovitskiy2021an} excel at extracting rich visual features, while CLIP\cite{radford2021learningtransferablevisualmodels} has revolutionized cross-modal retrieval by learning aligned representations of images and text through contrastive learning. In the audio domain, models like Wav2Vec 2.0\cite{NEURIPS2020_92d1e1eb} specialize in learning robust representations directly from raw waveforms, while CLAP\cite{10095969} enables sophisticated audio search through audio-text alignment. Video content presents additional challenges due to its temporal nature, addressed by models like ViViT\cite{9710415} and VideoPrism\cite{zhao2024videoprism} that capture both spatial and temporal dynamics in compact feature vectors. These various approaches enable efficient cross-modal search and retrieval, supporting applications from visual question-answering to temporal video localization.

The effectiveness of multimodal RAG systems fundamentally depends on these representation and retrieval capabilities. While significant progress has been made with text embedding models like BGE\cite{li2024makingtextembeddersfewshot} and NV-Embed\cite{lee2024nv}, extending these capabilities to handle multimodal data remains challenging. Beyond the core modalities discussed, emerging data types such as code, structured data, and graph-structured information require specialized representation and retrieval approaches. The development of more sophisticated cross-modal retrieval techniques represents a crucial direction for advancing multimodal RAG systems.

\subsubsection{Multimodal Understanding and Generation}
Building on effective multimodal representation and retrieval, multimodal understanding and generation are essential for advancing RAG systems beyond text-only processing. While representation and retrieval enable access to relevant multimodal content, the system must also process this information to comprehend cross-modal relationships and generate coherent outputs. Traditional RAG systems, designed primarily for textual data, often struggle to handle and generate information across modalities, leading to incomplete or inaccurate responses when multimodal content is crucial. Modern multimodal RAG systems address these limitations through advanced integration of understanding and generation across diverse modalities.

Models like CLIP \cite{radford2021learningtransferablevisualmodels} have pioneered the alignment of vision and language by embedding both data modalities into a shared space, thereby enabling effective and efficient cross-modal retrieval. Building upon this foundational work, systems such as MuRAG \cite{chen2022muragmultimodalretrievalaugmentedgenerator} further enhance question-answering capabilities by retrieving both images and text, while RA-CM3 \cite{yasunaga2023retrievalaugmentedmultimodallanguagemodeling} extends this functionality by enabling the simultaneous retrieval and generation of both text and images. Recent advancements such as Transfusion \cite{zhou2024transfusionpredicttokendiffuse} demonstrate the promising potential of unifying language modeling with diffusion, enabling the training of a single transformer model capable of both understanding and generating content across multiple modalities. Similarly, Show-o \cite{xie2024showosingletransformerunify} introduces a unified transformer architecture that combines autoregressive and discrete diffusion modeling techniques to flexibly support a broad range of vision-language tasks, including visual question-answering, text-to-image generation, and text-guided inpainting/extrapolation. Further innovations, like VisRAG \cite{yu2024visragvisionbasedretrievalaugmentedgeneration}, underscore the growing potential of multimodal RAG by leveraging both visual and textual information to achieve more comprehensive and nuanced document understanding, while LA-RAG \cite{li2024laragenhancingllmbasedasraccuracy} advances the field of speech processing by incorporating fine-grained, token-level speech retrieval mechanisms.

\subsection{Memory RAG}

Traditional RAG systems primarily rely on two extremes of knowledge storage and access: implicit memory embedded in language models through pre-training, and working memory that directly retrieves and processes raw text chunks. While language models excel at storing compressed general knowledge in their parameters, and traditional RAG effectively handles immediate context through text retrieval, there exists a crucial gap between these two approaches. Memory RAG bridges this gap by introducing explicit memory mechanisms that serve as an intermediate layer. This approach is particularly valuable for scenarios requiring comprehensive understanding of long documents or maintaining personalized knowledge over time, such as book comprehension and technical documentation analysis, where it can maintain hierarchical representations of document structure and key insights, as well as in personalization scenarios where it efficiently stores and updates compressed representations of user preferences and behavior patterns.

The value of this intermediate approach lies in its unique advantages over both extremes. Unlike the rigid implicit memory in language models that requires costly retraining to update, and unlike the computationally intensive working memory that processes raw text directly, explicit memory in Memory RAG provides a flexible and efficient middle ground. It enables compressed, structured representations of knowledge that are both more updateable than model parameters and more efficient than raw text processing. By implementing these distinct memory mechanisms, Memory RAG creates a more complete knowledge processing pipeline that combines the strengths of all three memory types. This section first explores these different types of memory and their complementary roles, then demonstrates how their technical implementation contributes to more efficient and scalable information processing in RAG systems.

\subsubsection{Types of Memory}

\begin{table}[t]
\caption{Comparison of implicit, explicit, and working memory in RAG systems across storage form, temporal duration, knowledge scope, usage frequency, update cost, and read cost.}
\label{tab:memory-types}
\centering
\begin{tabular}{cccc}
\toprule
\textbf{Aspects} & \textbf{Implicit Memory} & \textbf{Explicit Memory} & \textbf{Working Memory} \\
\midrule
Storage Form & model parameter & KV cache & plain text \\
Memory Term & long-term & long-term & short-term \\
Knowledge Scope & universal & domain-specific, personalized & task-specific, contextual \\
Usage Frequency & high & medium & low \\
Update Cost & high & low & low \\
Read Cost & low & low & high \\
\bottomrule
\end{tabular}
\end{table}

\textit{Implicit Memory.}
Implicit memory, analogous to human implicit memory \cite{gabrieli1988impaired,corkin2002s,bayley2005failure}, represents knowledge that is deeply embedded but not consciously accessible. Just as humans develop unconscious skills like riding a bicycle or typing without conscious thought, RAG systems encode implicit knowledge within their model parameters through training. This form of memory manifests through the pre-trained weights of both retriever and generator components, storing global knowledge and domain-specific patterns that are frequently accessed during inference. While implicit memory enables fast inference and provides a foundation for general knowledge, it shares limitations with its human counterpart: it's costly to acquire, difficult to modify, and not directly interpretable. Updates to this type of memory typically require model retraining.

\textit{Explicit Memory.}
Explicit memory parallels human explicit memory \cite{parkin1990differential}, the conscious recollection of facts and experiences. In RAG systems, it serves as a compressed, structured representation of long-term knowledge capturing high-level understanding of information sources. Unlike simple document retrieval, explicit memory stores abstract representations such as understanding of entire books, user behavior patterns, or structured summaries of knowledge bases. This intermediate form between raw text and model parameters enables efficient storage while maintaining semantic richness. It operates at medium access frequency, supporting stable yet updatable knowledge representation. The key advantage of explicit memory lies in its ability to preserve holistic understanding unattainable through simple passage retrieval while remaining more flexible than implicit memory encoded in model parameters.

\textit{Working Memory.}
Working memory mirrors human short-term memory \cite{cowan2001magical,cowan2012working}, serving as a temporary storage for immediate processing. In RAG systems, this manifests as the retrieved plain text chunks that are incorporated into the prompt context window, along with recent conversation history and intermediate computation results. Just as humans maintain recent information in mind while performing complex tasks, RAG's working memory temporarily holds relevant retrieved passages to inform the current generation task. This type of memory is characterized by high-frequency access, rapid updates with each new task, and temporary storage. It faces similar capacity constraints to human working memory - it's limited by computational resources and context window size, just as humans can only hold a finite amount of information in immediate awareness. The retrieved passages in the prompt serve as the temporary workspace where the model processes and synthesizes information to generate appropriate responses.

\subsubsection{Technical Implementation}
Memory RAG primarily focuses on explicit memory, which takes the form of sparse key-value caches that form an intermediate representation between raw text and model parameters. Memory RAG converts knowledge from raw text into more compact and structured explicit memories via key-value caching, using them to guide and optimize retrieval while reducing dependence on real-time search to improve efficiency. It reduces computational costs through efficient memory management, enhances long-text processing capabilities by storing and selectively retrieving information, and improves answer accuracy through structured knowledge representation.

$\text{Memory}^3$ \cite{yang2024memory3} employs a progressive compression mechanism that transforms raw input tokens into explicit memories through a two-stage pre-training process. The first stage, called warmup stage, focuses on memory formation without explicit memory to develop the model's basic reading comprehension abilities. The second stage, called continual train stage, introduces explicit memories and enhances the model's ability to utilize these memories during generation. This approach particularly emphasizes memory sparsification and storage optimization through sparse key-value storage, with several innovative designs: it selects only the first half of attention layers as memory layers, reduces key-value heads through grouped query attention, and employs token-level sparsification to select only the most relevant tokens for each head. These optimizations enable efficient handling of large-scale knowledge bases while significantly reducing storage requirements.

MemoRAG \cite{MemoRAG2024Hongjin} takes a unique approach by employing a lightweight but long-range LLM as a global memory system. This design adopts a dual-system architecture: a lightweight LLM serves as the memory module to form global memory of the database, while a more expressive LLM acts as the generation module. To form the global memory, MemoRAG introduces a progressive memory formation process where the memory module processes the input sequence through multiple attention layers. During this process, it maintains a KV cache of previous memory tokens and employs special weight matrices for memory formation, allowing the model to transform raw input tokens into compressed memory representations. The memory tokens are updated through attentive interactions with both the input context and previously formed memories, enabling the model to capture and preserve essential information across the entire database. The memory module first generates draft answers that serve as retrieval clues, effectively reducing the search space and improving retrieval precision. This approach is particularly effective for handling ambiguous information needs, distributed evidence gathering, and information aggregation tasks.

CAG (Cache-Augmented Generation) \cite{chan2024don} takes a radical approach by eliminating real-time retrieval from its workflow. Instead of performing retrieval during generation, it pre-computes key-value caches for all relevant documents and uses them directly in generation. The system operates in three phases: external knowledge preloading, where all relevant documents are preprocessed and formatted into a KV cache; inference, where the precomputed cache is used directly for generation without retrieval; and cache reset, which efficiently maintains system performance across multiple conversation turns. This design is particularly effective for scenarios with well-defined, manageable knowledge bases, where the complete set of relevant information can be pre-processed and stored in cache. The direct cache access strategy not only improves response time but also ensures stable performance in multi-turn interactions, while significantly reducing system complexity by removing the need to integrate separate retrieval and generation components.

HawkRAG~\cite{qian2025memory} extends the dual-system framework with a memory-enhanced retrieval approach for unstructured, long-context scenarios. It employs a lightweight module to build compressed KV memory representations of the context and generate draft answers for focused retrieval. A more expressive model then produces the final output from this refined context. It also introduces retrieval learning from generation feedback (RLGF) to enhance memory relevance by optimizing generation quality, achieving strong performance where traditional RAG methods struggle.

\begin{figure*}
    \centering
    \includegraphics[width=.85\linewidth]{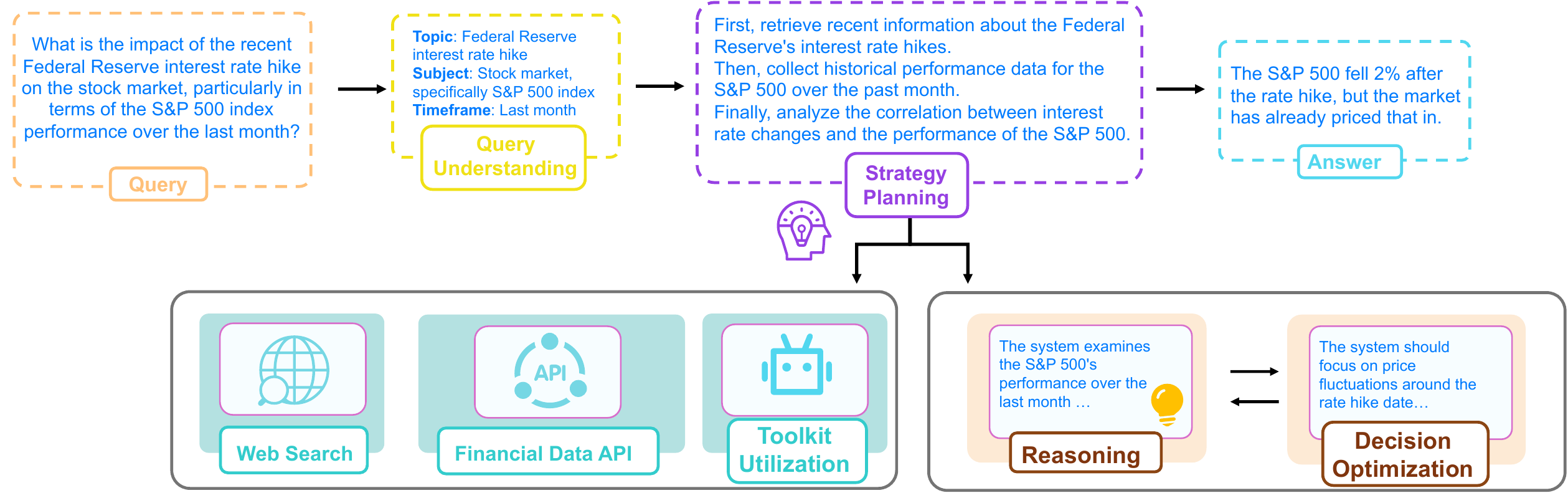}
    \caption{Agentic RAG consists of query understanding \& strategy planning, toolkit utilization, reasoning \& decision optimization.}
    \label{fig:Agentic_RAG}
\end{figure*}

\subsection{Agentic RAG}
Agentic RAG is an advanced framework integrating autonomous agents with RAG technology, significantly enhancing information retrieval and generation. This approach introduces agent-based decision-making to dynamically manage retrieval strategies, improving complex problem-solving, knowledge management, and generation. In Agentic RAG, agents handle query understanding, tool use, and reasoning optimization.

\subsubsection{Query Understanding \& Strategy Planning}

Query understanding and strategy planning is a critical step in Agentic RAG, enabling the effective understanding of user queries and the formulation of appropriate retrieval strategies. Agents analyze query complexity and topics to determine the priority and methods of retrieval. For instance, AT-RAG\cite{at_rag_2024} improves retrieval efficiency and reasoning accuracy for complex multi-hop queries through topic filtering and iterative reasoning. Additionally, REAPER \cite{reaper_2024} introduces a reasoning-based retrieval planner, enhancing performance in complex queries, which demonstrate how query understanding can provide more efficient retrieval pathways for Agentic RAG.

\subsubsection{Toolkit Utilization}

A key feature of Agentic RAG is its ability to utilize multiple tools. Agents are capable of employing traditional search engines, calculators, APIs, and other external tools to enhance retrieval and reasoning capabilities. For example, AT-RAG \cite{at_rag_2024} integrates various tools to expand the scope of retrieval, while RAGENTIC\cite{ragentic_2024} leverages multi-agent collaboration and tool usage to tackle complex tasks. This integration of multiple tools allows Agentic RAG to adapt flexibly to different task requirements, thus improving the system's efficiency and adaptability.

\subsubsection{Reasoning \& Decision Optimization}
Reasoning and decision optimization is a core mechanism in Agentic RAG systems, ensuring that agents make optimal decisions based on constantly changing information and environments. Through reasoning, agents can evaluate the reliability of multiple information sources, perform multi-step reasoning, and optimize retrieval strategies based on reasoning outcomes. PlanRAG \cite{lee-etal-2024-planrag} improves the performance of generative LLMs in decision-making by employing a plan-then-retrieve approach. Moreover, REAPER \cite{reaper_2024} excels in decision optimization by refining retrieval planning through reasoning, thereby improving the response speed. These decision optimization methods not only enhance retrieval accuracy but also strengthen the ability to handle complex tasks.

\subsection{Parametric RAG}

Parametric RAG introduces a new paradigm that integrates retrieved knowledge directly into the parameters of LLMs, overcoming the inherent limitations of in-context knowledge injection. Unlike conventional RAG methods that append documents into the input prompt, Parametric RAG parameterizes each document into a set of low-rank matrices that can be merged into the feed-forward layers of the model, enabling the LLM to utilize external knowledge in the same manner as its internal parametric knowledge\cite{10.1145/3726302.3729957}. This parameter-level integration significantly reduces online computation and enhances reasoning efficiency in knowledge-intensive tasks.

\subsubsection{Offline Document Parameterization}

The offline parameterization process converts each document into a lightweight parametric representation that can later be plugged into an LLM’s parameters. As described by \cite{10.1145/3726302.3729957}, this involves two main steps: (1) \textit{Document Augmentation}, which enriches each document via rewriting and question-answer (QA) generation to capture factual diversity; and (2) \textit{Parametric Encoding}, which trains low-rank matrices (\(A,B\)) following the LoRA method\cite{hu2022lora} to embed document knowledge within the model’s feed-forward layers. These LoRA parameters (\(\Delta\theta = A B^\top\)) serve as a document-specific representation that can be merged efficiently at inference time. Empirical results show that this parameterization allows models to internalize document-level knowledge more effectively than prompt concatenation, while keeping computational cost affordable through offline precomputation.

\subsubsection{Retrieve-Update-Generate Workflow}

During inference, Parametric RAG follows a \textit{Retrieve-Update-Generate} (RUG) workflow\cite{10.1145/3726302.3729957}. First, the retriever identifies the top-$k$ relevant documents, each already associated with its parametric representation. Second, these representations are merged into the model via low-rank addition (\(\Delta W_{merge} = \alpha \sum_j A_j B_j^\top\)), forming an updated model \(L'(\theta')\) that integrates the retrieved knowledge temporarily within its parameters. Finally, the model generates responses directly based on the query without adding text tokens into the context. This workflow eliminates long-context overhead and improves reasoning accuracy and response latency. Comparative experiments on benchmarks such as HotpotQA, 2WikiMultihopQA, and ComplexWebQuestions confirm that Parametric RAG reduces inference time by up to 36\% while maintaining or exceeding in-context RAG baselines.

\subsubsection{Dynamic Parameter Integration and Hybrid Models}

Building on the foundational framework of Parametric RAG, \cite{10.1145/3726302.3731692} propose \textit{Dynamic and Parametric RAG}, which dynamically balances between in-context and parametric knowledge integration based on retrieval confidence and query complexity. This dynamic mechanism allows the system to switch adaptively between retrieval at the prompt level and parameter-level updates, optimizing the trade-off between computational efficiency and reasoning flexibility. The hybrid approach demonstrates that parameter-level knowledge injection and in-context retrieval are complementary rather than conflicting paradigms. Moreover, this dynamic extension paves the way for scalable, continually updatable knowledge systems, bridging RAG and fine-tuning paradigms for future large-scale applications.

\section{Comprehensive Evaluation Strategies for RAG}
\label{sec:eval}
Evaluating RAG systems necessitates a comprehensive assessment of both their effectiveness and efficiency. Effectiveness evaluations focus on how well a model retrieves the most relevant context, integrates it into coherent and accurate answers, and preserves overall consistency. Efficiency assessments, on the other hand, consider the computational resources required, the response latency, and the scalability of the retrieval and generation processes.

\subsection{Assessing the Effectiveness of RAG}
Evaluating the effectiveness of RAG is key in determining their capability to harness external knowledge for generating reliable responses. The following subsections delve into specific metrics designed to evaluate key aspects of RAG effectiveness, including query-context relevance, context-answer coherence, and query-answer accuracy.

\subsubsection{Assessing Query-Context Relevance}
In RAG systems, the ability to retrieve pertinent information in response to user queries is fundamental to providing large language models (LLMs) with the necessary context for accurate reasoning and generation~\cite{cheng2024towards}. This process involves efficiently integrating data from diverse knowledge sources and implementing robust filtering mechanisms to extract relevant information while minimizing noise from less reliable or extraneous data. To evaluate the effectiveness of RAG systems in this regard, several benchmarks and metrics have been developed. Multiple frameworks including ARES\cite{saadfalcon2024aresautomatedevaluationframework}, RAGAS\cite{es2023ragasautomatedevaluationretrieval}, and TruLens\cite{trulens} employ Context Relevance metrics to assess how well the retrieved context aligns with and supports the user's query. Additionally, the KILT\cite{petroni2021kiltbenchmarkknowledgeintensive} benchmark, designed for knowledge-intensive tasks, offers metrics such as R-precision and Recall@k to evaluate the quality of the retrieved context, emphasizing its relevance and comprehensiveness concerning the original query. While these benchmarks share a common goal of ensuring that the retrieved context meaningfully supports the user's query, they differ in their specific approaches and emphases. ARES, RAGAS, and TruLens focus on the immediate alignment between query and context, whereas KILT provides a broader evaluation by considering the richness and completeness of the retrieved information. Collectively, these metrics are critical for assessing a RAG system's ability to retrieve and provide high-quality, relevant information, enabling LLMs to generate accurate and contextually grounded responses. However, a key limitation of these metrics is their reliance on semantic similarity, which may not always capture true relevance; a document can be topically similar yet fail to contain the specific facts needed to answer the query. Furthermore, automated metrics often struggle to penalize "near-miss" retrievals effectively.

\subsubsection{Evaluating Context-Answer Coherence}
The relationship between retrieved context and generated answers is crucial for ensuring response quality in RAG systems. The generator must produce responses that are coherent and faithful to the retrieved context, minimizing hallucinations and unsupported claims. To evaluate this capability, several benchmarks and metrics have been developed. For instance, CRUD~\cite{lyu2024crudragcomprehensivechinesebenchmark} employs Recall and Precision to assess the accuracy of generated answers, emphasizing correctness and completeness. ARES and RAGAS introduce the Answer Faithfulness metric, measuring how closely generated answers adhere to retrieved passages, ensuring content is free from hallucinations. RECALL~\cite{liu2023recallbenchmarkllmsrobustness} proposes two key metrics: Accuracy, assessing factual correctness, and Misleading Rate (M-Rate), evaluating the frequency of misleading information. TruLens introduces Groundedness, which decomposes the LLM’s output into individual claims and verifies each against retrieved context to ensure evidential support. While these benchmarks share the goal of evaluating the fidelity and coherence of generated answers, they differ in focus: CRUD measures Recall and Precision; ARES and RAGAS emphasize faithfulness; RECALL evaluates accuracy and misleading tendencies; and TruLens delves into the granular verification of individual claims.

In addition to these core metrics, recent work has focused on enhancing the reliability of RAG systems through high-quality citation and accurate attribution\cite{zhou2024trustworthinessretrievalaugmentedgenerationsystems}. For instance, LongBench-Cite\cite{zhang2024longciteenablingllmsgenerate} is designed to evaluate models in Long-Context Question Answering with Citations (LQAC), where the relevance and accuracy of citations are essential for maintaining transparency. The RECLAIM framework\cite{xia2024groundsentenceimprovingretrievalaugmented} further develops this idea with the Correct Attribution Score (CAS), assessing whether generated answers are fully supported by citations, and the Citation Redundancy Score (CRS), which minimizes unnecessary citations to enhance response clarity. Metrics such as Verifiability and Consistency Ratio (CR) in RECLAIM ensure that the retrieved data genuinely supports the generated content.

Moreover, RARR~\cite{gao2023rarrresearchingrevisinglanguage} introduces sentence-level AIS and Levenshtein distance to measure citation accuracy, ensuring that references retain their original meaning. WebGPT~\cite{nakano2022webgptbrowserassistedquestionansweringhuman} adds a layer of reliability by distinguishing between imitative and non-imitative errors to improve error correction. Additionally, ALCE~\cite{gao2023enablinglargelanguagemodels} uses citation recall and precision to evaluate citation relevance and conciseness. Collectively, these metrics contribute to a comprehensive framework for evaluating RAG's ability to generate contextually grounded, accurate, and transparently attributed responses, emphasizing reliability and trustworthiness in practical applications. Despite their utility, these metrics face challenges. Automated faithfulness checks (like in RAGAS or ARES) can be susceptible to surface-level agreement, potentially missing subtle hallucinations or contradictions. Conversely, metrics requiring granular claim verification are more robust but introduce significant computational overhead, making them difficult to apply at scale.

\subsubsection{Measuring Query-Answer Accuracy}
In RAG systems, accurately measuring the alignment between user queries and generated answers is essential for evaluating overall performance. Traditional QA evaluations often rely on accuracy metrics, which assess whether the system provides correct answers. However, RAG systems introduce additional complexities, requiring more nuanced evaluation methods. A key challenge in RAG evaluation is hallucination, instances where the model generates content not grounded in retrieved context. To address this, benchmarks such as CRAG~\cite{yang2024cragcomprehensiverag} incorporate metrics like hallucination rate and missing rate alongside accuracy, calculating their difference to penalize hallucinations and prioritize factual correctness. Other frameworks, including RAGAS, ARES, and TruLens, introduce an Answer Relevance metric to evaluate how well the generated response aligns with the query. Similarly, CRUD proposes RAGQuestEval to assess answer relevance and correctness, while KILT evaluates RAG systems on specific downstream tasks, providing a task-oriented perspective. The RGB framework~\cite{chen2023benchmarkinglargelanguagemodels} further extends evaluation with rejection, error detection, and correction rates to test a system’s ability to identify and reject misleading information. Collectively, these approaches reflect a shift toward comprehensive RAG evaluation that emphasizes relevance, factual accuracy, and error handling over simple accuracy. However, a major limitation is the reliance on automated LLM-as-a-judge evaluations (common in RAGAS, ARES, and CRAG), which may inherit the biases and inaccuracies of the judging model. While more objective than raw accuracy, such metrics remain imperfect substitutes for human evaluation, particularly for nuanced, domain-specific, or multi-faceted queries.

\subsection{Evaluating the Efficiency of RAG}

Evaluating the efficiency of RAG systems involves analyzing how retrieval and generation jointly affect performance. Efficiency depends on latency, throughput, and resource utilization, which vary across paradigms such as Agentic, Graph, Memory, and Parametric RAG. Understanding these trade-offs is key to designing low-cost, high-performance systems suitable for real-world deployment.

\subsubsection{Efficiency Metrics and Benchmarks}
As RAG systems progress from research prototypes to industrial deployment, efficiency has become a key factor for scalability and usability. Because RAG couples retrieval with generation, overall performance depends on the latency, throughput, and resource utilization of both components. Emerging paradigms, such as Agentic RAG, Graph RAG, Memory RAG, and Parametric RAG, offer different trade-offs between reasoning flexibility and computational cost. Evaluating these differences and developing low-cost, high-efficiency architectures are key to enabling RAG systems that are both practically deployable and economically sustainable.

Several benchmarking frameworks have been developed to quantify these metrics. The RAG-Performance library \cite{SciPhiAI_RAGPerformance} simulates high-volume workloads to assess system throughput and responsiveness under stress, offering a unified benchmark for both retrieval and generation latency. Other studies expand efficiency analysis to include hardware-level profiling, identifying trade-offs between computational effectiveness and resource consumption \cite{csakarmaximizing,jin2024ragcacheefficientknowledgecaching,feng2024easyragefficientretrievalaugmentedgeneration}. These tools form a solid foundation for understanding operational efficiency across various configurations and workloads.

\subsubsection{Efficiency across RAG Paradigms}
Beyond standard pipeline-based architectures, different RAG paradigms exhibit distinct efficiency characteristics. Agentic RAG structures retrieval and generation as iterative reasoning loops, where an autonomous agent dynamically decides when and what to retrieve. This adaptivity improves factual accuracy but often increases latency due to repeated retrieval-generation cycles and decision-making overhead. Graph RAG, in contrast, represents knowledge as structured entities and relations, enabling graph traversal instead of redundant retrievals, thereby improving caching, parallelism, and scalability. Memory RAG integrates long-term memory components that cache previous retrievals or conversation histories, reducing redundant queries in multi-turn or persistent contexts, though memory synchronization and freshness maintenance can become bottlenecks. Meanwhile, Parametric RAG embeds part of the external knowledge directly within model parameters, reducing retrieval frequency and improving latency but increasing training cost and reducing adaptability when knowledge updates are frequent. Comparing these paradigms reveals a key trade-off between reasoning adaptivity, computational cost, and latency efficiency, suggesting that future evaluation frameworks should consider paradigm-level efficiency beyond model-level performance.

\subsubsection{Toward Low-cost and Scalable RAG}
Constructing low-cost, high-efficiency RAG systems requires end-to-end optimization across the retrieval-generation pipeline. Promising directions include adaptive retrieval strategies that dynamically adjust retrieval depth based on query complexity, hierarchical indexing for faster evidence selection, and caching or distillation mechanisms, such as RAGCache \cite{jin2024ragcacheefficientknowledgecaching}, that reuse prior retrievals to minimize redundant computation. Lightweight retrieval-generation coordination frameworks can further reduce communication overhead, improving throughput without compromising accuracy. Future research should also integrate paradigm-aware efficiency benchmarks that jointly evaluate computational cost, reasoning efficiency, and retrieval redundancy. These advancements are critical for bridging the gap between academic research and industrial deployment, ensuring that RAG systems remain not only effective but also economically sustainable at scale.

\section{Downstream Tasks and Applications}

This section reviews how RAG is applied to downstream tasks and domain-specific scenarios. By coupling parametric language models with external knowledge retrieval, RAG provides a flexible framework for tasks that require factual grounding, contextual enrichment, and up-to-date information access. We first summarize representative downstream tasks, including question answering, information extraction, and text understanding, and then discuss how RAG has been adapted to practical applications across scientific, medical, legal, educational, and industrial domains.
\subsection{Downstream Tasks}

\begin{table}[t]
\caption{Representative datasets used to evaluate RAG systems across downstream tasks, including question answering, information extraction, and text understanding, together with their corresponding sub-task categories.}
\label{tab:datasets}
\centering
\begin{tabular}{c|l|l}
\hline
\textbf{Task} & \textbf{Sub Task} & \textbf{Dataset} \\ \hline
\multirow{14}{*}{Question Answering (QA)} & \multirow{6}{*}{Single-hop QA} &
Natural Question (NQ) \cite{kwiatkowski2019natural}   \\
 &  & TriviaQA (TQA) \cite{joshi2017triviaqa}  \\
 &  & SQuAD \cite{rajpurkar2016squad}   \\
 &  & Web Questions (WebQ) \cite{berant2013semantic}  \\
 &  & PopQA~\cite{mallen2022not}  \\
 &  & CRAG \cite{yang2024crag}   \\   \cline{2-3}
 & \multirow{5}{*}{Multi-hop QA} & HotpotQA \cite{yang2018hotpotqa}    \\
 &  & 2WikiMultiHopQA \cite{ho2020constructing}  \\
 &  & StrategyQA \cite{geva2021did}     \\
 &  & MuSiQue \cite{trivedi2022musique}         \\
 &  & MultiHop-RAG  \cite{tang2024multihop}   \\ \cline{2-3}
 & \multirow{3}{*}{Long-form QA} & ELI5 \cite{fan2019eli5}  \\
 &  & WebCPM \cite{qin2023webcpm}  \\
 &  & NarrativeQA (NQA) \cite{kovcisky2018narrativeqa} \\  \cline{2-3}  \hline
\multirow{4}{*}{Information Extraction} & \multirow{2}{*}{Entity Linking} &  ZESHEL \cite{logeswaran2019zero}     \\
 &  & CoNLL \cite{hoffart2011robust}   \\ \cline{2-3}
 & \multirow{2}{*}{Relation Extraction}  & T-REx \cite{elsahar2018t}      \\
 &  & ZsRE \cite{levy2017zero}    \\  \hline
 \multirow{6}{*}{Text Understanding} & \multirow{2}{*}{Text Classification}  & TREC \cite{li2002learning}  \\
 &  & SST-2 \cite{socher2013recursive} \\ \cline{2-3}
 &  \multirow{2}{*}{Text Summarization} &   WikiASP \cite{hayashi2021wikiasp}  \\
 &  & XSum \cite{narayan2018don}  \\ \cline{2-3}
 &  Text Generation &   Biography \cite{lebret2016neural}  \\ \cline{2-3}
 \hline
\end{tabular}
\end{table}

Generally, downstream tasks cover diverse areas, including question answering, information extraction, text generation, and classification. Table~\ref{tab:datasets} summarizes each task type and their common sub-tasks, highlighting areas where RAG models and similar architectures are frequently applied.

In question answering (QA), RAG models demonstrate significant advantages by retrieving targeted information that supports accurate and relevant answers. For straightforward factual questions, single-hop QA typically requires only one document or passage that directly addresses the query, allowing the model to generate precise answers with minimal reasoning complexity. In multi-hop QA, RAG gathers and synthesizes evidence from multiple sources to answer questions that connect different facts or viewpoints, enabling the model to handle more complex reasoning chains and deliver more in-depth responses. For long-form QA, where detailed paragraph-level answers are required, RAG can retrieve and combine multiple relevant documents to produce comprehensive, coherent, and well-informed responses, especially for open-ended or complex questions.

In information extraction, RAG enhances the accuracy and depth of entity and relation identification by pulling contextually relevant evidence from external sources. For entity linking, retrieved information about specific entities helps the model map mentions to standardized entries in a knowledge base, improving disambiguation and contextual understanding. For relation extraction, retrieval provides background information about related entities, supporting more accurate relationship identification and facilitating the construction of structured knowledge resources such as knowledge graphs.

In text understanding and generation, RAG provides contextual depth by retrieving supplementary information that improves interpretability and relevance. For text classification, especially in specialized domains, retrieved background knowledge or examples can support more accurate categorization. For text summarization, related documents or contextual evidence help the model focus on the most salient points and produce concise, informative summaries. For text generation, retrieval enables the model to incorporate timely data, domain-specific knowledge, or additional context into generated content.

\subsection{Applications of RAG}
RAG models have broad applications across various fields, where they combine the strengths of information retrieval with language generation. This part explores key application areas where RAG models significantly enhance performance by providing accurate, contextually relevant, and real-time information retrieval capabilities.

\subsubsection{AI for Science}
In scientific research, RAG models support researchers by enabling access to and integration of up-to-date information from extensive scientific databases, enhancing productivity and accelerating discovery.
For instance, in materials science, large language models like GPT-4 combined with retrieval capabilities can autonomously assist in complex tasks such as material design and data analysis, reshaping the research paradigm in this interdisciplinary field~\cite{yu2024large}.
In chemistry, RAG models improve discipline-specific query handling, supporting synthesis planning, reaction prediction, and understanding molecular interactions that require precise information beyond pre-trained knowledge~\cite{wellawatte2024chemlit}.
In physics, RAG models enhance question answering and comprehension by retrieving relevant context, allowing researchers and students to engage more deeply with complex physical concepts and supporting advanced inquiry and problem-solving~\cite{anand2023sciphyrag}.
In the life sciences, BioRAG~\cite{wang2024biorag} serves as a specialized question-answering system, providing efficient access to relevant biological information.
These examples underscore the versatility of RAG models across scientific domains, improving the retrieval and synthesis of domain-specific knowledge.

\subsubsection{Deep Research}
Beyond standard retrieval pipelines, DeepResearch-style agents combine long-horizon planning, web-scale retrieval, and iterative verification to conduct end-to-end investigations.
WebWalker establishes a benchmark for web traversal, emphasizing multi-hop navigation, tool use, and grounding, thus providing a principled way to quantify the deep-research capabilities of LLM agents~\cite{webwalker2025}.
Building on this, WebDancer introduces an autonomous information-seeking framework coupling planner-executor loops with adaptive querying and reflection, enabling agents to decompose open-ended questions and refine evidence across sessions~\cite{webdancer2025}.
To enhance verifiability and synthesis at scale, WebWeaver structures retrieved evidence with dynamic outlines, organizing heterogeneous web content into hierarchical artifacts that support faithful aggregation and citation~\cite{webweaver2025}.
Together, these systems illustrate a trajectory for deep research: from measurable web navigation, to agentic information seeking, to structured, verifiable knowledge assembly, demonstrating how RAG underpins long-horizon, trustworthy research agents.

\subsubsection{Finance}
In the financial domain, RAG models play a crucial role in enhancing the accuracy and relevance of information derived from complex data, supporting tasks like information extraction, question answering, and sentiment analysis.
In question answering on financial documents, RAG models enhance response reliability by refining retrieval methods. Techniques like improved chunking and query expansion ensure that the model retrieves the most relevant information, resulting in more accurate and contextually appropriate answers in financial QA systems \cite{setty2024improving}.
For financial sentiment analysis, RAG models offer substantial advantages by retrieving additional context, allowing large language models to make more accurate sentiment predictions. This capability aligns model outputs with the nuanced requirements of financial sentiment, supporting better-informed valuation and investment decisions \cite{zhang2023enhancing}.
Finally, RAG-augmented models provide valuable business insights by extracting data from various document types, such as invoices and reports. This approach offers organizations a scalable and cost-effective solution for comprehensive data analysis and insight generation \cite{arslan2024business1, arslan2024business2}.
These applications demonstrate how RAG models address specific challenges in finance, improving information retrieval and supporting more informed decision-making across the industry.

\subsubsection{Education}

In education and e-learning, RAG models enhance learning experiences by providing personalized, context-specific support.
Intelligent tutoring systems using RAG overcome traditional limitations by delivering accurate, interactive assistance that improves student engagement and academic performance~\cite{modran2024llm, thus2024exploring}.
Additionally, HiTA is a RAG-based platform that supports educators by positioning them within AI-assisted learning loops, functioning as a teaching assistant that increases course-specific learning satisfaction~\cite{liu2024hita}.
Furthermore, Education-Specific RAG AI (ES-RAG AI) ensures transparency and accountability in AI-generated educational content, aligning it with educational needs~\cite{elmessiry2024navigating}.
Overall, RAG models create adaptive learning environments, tailor content based on student progress, and assist educators in efficient curriculum development, enhancing both learning and teaching processes.

\subsubsection{Healthcare}

In healthcare, RAG has emerged as a transformative tool, enhancing how medical professionals access and utilize medical knowledge. By combining advanced retrieval mechanisms with generative capabilities, RAG significantly supports knowledge retrieval, decision-making processes, and personalized healthcare services.
RAG applications span diverse areas, such as Traditional Chinese Medicine, where it retrieves specialized knowledge from texts like the Compendium of Materia Medica, aiding modernization efforts~\cite{liu2024application}.
In infectious disease management, RAG-based chatbots provide timely, context-aware information to address healthcare resource shortages~\cite{raja2024rag}.
In precision medicine, RAG integrates NLP and genomic data to deliver personalized treatment recommendations for conditions like Multiple Myeloma~\cite{quidwai2024rag}.
These examples illustrate RAG’s versatility in healthcare, supporting both traditional and modern medicine, improving disease management, and advancing precision medicine.

\subsubsection{Legal}

In the legal sector, RAG models help navigate complex legal texts, supporting legal research, document drafting, and client consultations.
For example, CBR-RAG combines Case-Based Reasoning with RAG to structure retrieval processes, enhancing legal question-answering by ensuring contextually relevant cases inform the LLM's outputs~\cite{wiratunga2024cbr}.
LegalBench-RAG introduces a benchmark emphasizing precise retrieval in the legal domain, focusing on extracting relevant text to support accurate analysis and mitigate context window limitations~\cite{pipitone2024legalbench}.
Additionally, HyPA-RAG, a hybrid parameter-adaptive system, addresses challenges such as retrieval errors and outdated information by dynamically adjusting retrieval parameters, significantly improving response accuracy in high-stakes applications like policy interpretation~\cite{kalra2024hypa}.  Furthermore, recent studies\cite{10.1145/3624918.3625321} have shown that the way legal case retrieval systems present results can substantially influence user decision-making, underscoring the importance of unbiased and transparent retrieval in sensitive domains like law.
These implementations highlight how RAG models enhance efficiency, accuracy, and contextual relevance in legal research and documentation, ultimately enabling faster, more reliable legal services.

\subsubsection{Industry}

In industry, RAG models play a vital role in improving operational efficiency, real-time troubleshooting, and decision-making in complex scenarios by integrating retrieval with generative capabilities.
For instance, RAG systems in industrial troubleshooting retrieve information from technical manuals, expert databases, and real-time sensor data, synthesizing it to offer context-specific solutions, thereby minimizing downtime and enhancing resolution rates~\cite{narimaniintegration}.
In industrial maintenance, STMicroelectronics used RAG as cognitive assistants to combine technical document retrieval with generative capabilities to support advanced maintenance strategies, reducing errors and providing context-rich assistance~\cite{machado2024development}.
Additionally, Intelligent Driving Assistance Systems use RAG models to provide real-time, context-based guidance from car manuals, significantly enhancing the driving experience~\cite{hernandez2024idas, kieu2024empowering, liu2024optimizing}.
Overall, RAG models bridge the gap between data retrieval and contextual generation in industrial applications, delivering timely insights that reduce operational challenges and improve efficiency across various sectors.

\section{Prospects and Future Directions}

\subsection{GraphRAG} \textbf{Significance.} With growing attention to knowledge-graph-based methods in retrieval-augmented generation (RAG) systems, effectively leveraging structured relationships for tighter interaction between text generation and retrieval has become a new research focus. By coupling retrieved external knowledge with graph structures, GraphRAG establishes explicit connections among entities and their relations, leading to more interpretable and robust performance, especially for complex factual or reasoning tasks. In practice, GraphRAG can also be viewed as a context extension of traditional RAG, where nodes and edges in the graph serve as composable contexts for multi-hop reasoning and entity alignment.

\textbf{Open issues.} Although some studies have integrated knowledge graphs into RAG, a systematic approach to incorporating multi-source, heterogeneous graph data across cross-domain or cross-lingual scenarios remains lacking. Furthermore, dynamically updating or correcting inaccurate or outdated graph structures during inference remains a major challenge. As the scope of application scenarios expands, real-time responsiveness on large-scale graphs, efficient retrieval, and scalability in high-concurrency settings are emerging as key directions for future GraphRAG research.

\subsection{Multimodal RAG}
\textbf{Significance.} Multimodal RAG is becoming increasingly prevalent, as real-world information is not limited to text but also includes images, videos, audio, and other sensor data. Effectively incorporating these heterogeneous modalities into RAG can significantly enhance reasoning capabilities, enabling more comprehensive understanding and generation in applications such as human-machine interaction, multimodal content creation, and medical diagnostics. By integrating visual, auditory, and textual information, Multimodal RAG can improve tasks like product recommendation, scene understanding, and multimodal QA, making retrieval-enhanced reasoning more robust and context-aware.

\textbf{Open Issues.} A major challenge in Multimodal RAG lies in how to effectively represent and retrieve information across different modalities, ensuring seamless fusion with large language models for reasoning and generation. Heterogeneous data sources often have distinct structures, resolutions, and levels of noise, requiring advanced alignment strategies to maintain consistency and interpretability. Moreover, real-time retrieval and reasoning over high-dimensional multimodal data demand efficient indexing, storage, and processing techniques. Future research should focus on scalable multimodal representation learning, cross-modal retrieval optimization, and adaptive fusion strategies, while also addressing issues such as privacy, robustness, and incremental updates in dynamic environments.

\subsection{Personalized RAG}
\textbf{Significance.} Personalized RAG is crucial for enhancing user experience in recommender systems, dialogue agents, and QA applications. By integrating user-specific knowledge, such as preferences and historical interactions, RAG can generate more relevant and context-aware responses. A practical and cost-effective approach to achieving personalized RAG is through memory-based mechanisms, where user-related knowledge is systematically extracted, stored, and retrieved. This enables the model to adapt dynamically to individual needs while maintaining efficiency and scalability.

\textbf{Open Issues.} The key challenges in Personalized RAG concern memory extraction, storage, knowledge updating, and fusion. Efficiently structuring and managing memory to ensure relevance and usability over time remains an open problem. Moreover, updating personalized knowledge while mitigating forgetting or inconsistencies requires advanced lifelong learning strategies. Privacy and security concerns persist, as using personal information must comply with data protection regulations and prevent unauthorized access. Extending Personalized RAG to multilingual and cross-domain scenarios further requires research to ensure adaptability across diverse linguistic and cultural contexts.

\subsection{Agentic RAG}
\textbf{Significance.}
Agentic RAG operates like an intelligent investigator. It not only retrieves relevant documents but also analyzes content, cross-references information, and simulates interviews for deeper insights. This paradigm introduces AI agents as intermediaries between users and knowledge sources, substantially enhancing retrieval-augmented reasoning. These agents interpret queries, plan retrieval, use tools such as search engines, calculators, and APIs, and adapt reasoning to optimize synthesis. Recent advancements such as OpenAI’s DeepResearch\footnote{https://openai.com/index/introducing-deep-research/} and Google’s intelligent retrieval systems\footnote{https://research.google/blog/accelerating-scientific-breakthroughs-with-an-ai-co-scientist/} highlight the potential of Agentic RAG. DeepResearch shows how agents autonomously decide when and how to search, engage in multi-round reasoning and retrieval cycles, and progressively analyze information to produce comprehensive reports. Integrating agentic decision-making grants RAG systems greater autonomy, enabling them to handle complex, multi-step tasks beyond traditional frameworks.

\textbf{Open Issues.}
Despite its promise, Agentic RAG encounters notable challenges in structured multi-step reasoning, including maintaining interpretability, mitigating error propagation, and coordinating diverse knowledge sources effectively. While systems like DeepResearch  excel at identifying task-relevant and high-quality data sources, they are currently constrained to publicly accessible information, potentially overlooking insights from closed or proprietary datasets. Another challenge lies in the depth of analysis: current models tend to perform better at summarization than providing nuanced analytical insights. Developing reliable agents that can flexibly plan retrieval strategies while ensuring efficiency and correctness remains an open research area. Moreover, optimizing these agents for distributed and hardware-heterogeneous environments is critical for broader real-world deployment. Future work should focus on enhancing autonomous decision-making, integrating more robust reasoning mechanisms, and improving adaptive learning capabilities. Collaborating with domain experts to facilitate deeper analytical insights and expanding the scope of accessible data sources are essential directions for further advancing the effectiveness of Agentic RAG.

\subsection{RAG and Generative Models}

\textbf{Significance.} The integration of RAG with various generative models, including diffusion models and other deep generative architectures, opens up new possibilities for solving complex tasks. Beyond the synergy with large language models like ChatGPT or GPT-4, the incorporation of generative paradigms such as diffusion models can enhance both the diversity and fidelity of generated content. This broader integration extends RAG’s impact beyond traditional QA and dialogue tasks, fostering innovative applications in cross-domain scenarios such as scientific literature mining, medical imaging synthesis, and multimodal content generation.

\textbf{Open Issues.} Despite its considerable potential, the integration of RAG with generative models remains a formidable challenge due to the inherent opacity of both retrieval-based and generative processes. The limited interpretability of interactions between external knowledge retrieval and generative components significantly complicates model transparency and control. A more cohesive and unified approach in designing architectures, training paradigms, and retrieval-augmentation strategies is vital for effectively leveraging these powerful capabilities. Furthermore, robust and well-established mechanisms for uncertainty quantification, error correction, and adaptive retrieval are indispensable for mitigating inaccuracies and ensuring the long-term reliability of generated content in real-world applications.

\subsection{EdgeRAG}
\textbf{Significance.} As edge computing gains prominence in applications such as IoT, mobile systems, and smart environments, the role of RAG in edge scenarios is becoming increasingly important. Deploying retrieval and generation capabilities at the edge, rather than relying solely on cloud-based processing, offers key benefits such as lower latency, reduced bandwidth usage, and enhanced local privacy protection. These advantages are particularly crucial for real-time applications and data-sensitive scenarios where offloading to the cloud is impractical or undesirable.

\textbf{Open Issues.} The resource constraints in edge environments, including limited computational power, storage, and caching capacity, pose significant challenges for efficient RAG deployment. Optimizing model compression, retrieval efficiency, and adaptive caching strategies is essential to balance performance and resource consumption. Additionally, maintaining reliability and consistency in RAG outputs under network instability or intermittent cloud connectivity remains an open problem. Given the security vulnerabilities of edge environments, robust defense mechanisms against adversarial attacks and unauthorized access are also critical for ensuring safe and trustworthy EdgeRAG applications.

\subsection{Trustworthy RAG}
\textbf{Significance.} As RAG systems are increasingly deployed in real-world applications, their interpretability and trustworthiness have become critical concerns. Trustworthy RAG emphasizes not only transparency in retrieval and generation but also explainability of underlying mechanisms and cited sources. Ensuring interpretable retrieval pathways and outputs is essential for validating accuracy, enhancing user trust, and supporting decision-making in high-stakes domains such as healthcare, finance, and law. Beyond interpretability and factual reliability, recent research also highlights security resilience as a key component of trust. In particular, \textit{knowledge poisoning}, malicious manipulation of external knowledge sources, poses an emerging risk. Studies show that even a few crafted or misleading documents injected into large-scale databases, including Wikipedia or enterprise repositories, can compromise generation fidelity, leading models to produce attacker-specified or biased outputs~\cite{liang-etal-2025-saferag,DBLP:journals/corr/abs-2402-07867}. As RAG systems rely on dynamically updated, user-contributed knowledge, safeguarding data provenance and retrieval integrity is central to maintaining trustworthy generation.

\textbf{Open Issues.} Despite growing attention to trust and explainability, no universally accepted framework exists for assessing RAG reliability. Current mechanisms struggle to address challenges such as unverifiable citations, hallucinations, and inconsistencies between retrieved knowledge and generated content. Recent work also shows that defenses against knowledge poisoning, such as paraphrasing, perplexity filtering, and retrieval consistency checks, remain ineffective, underscoring the need for proactive security auditing and provenance verification. Beyond mitigating hallucination and bias, future research should develop interpretable and robust RAG architectures that can authenticate external sources, detect abnormal retrieval patterns, and ensure alignment with ethical and regulatory standards. Only by integrating transparency, security, and verifiability can RAG systems achieve genuine trustworthiness in real-world deployments.

\subsection{RAG vs. Long-context LLMs}
\textbf{Significance.}
With the rapid expansion of context windows in large language models (LLMs), modern systems such as GPT-4 Turbo, Gemini 1.5 \cite{geminiteam2024gemini15unlockingmultimodal}, and Claude 3 can process hundreds of thousands, or even millions, of tokens within a single prompt. This development raises a fundamental question: in which scenarios does RAG still offer clear advantages over long-context LLMs? While long-context models enable direct reasoning over large volumes of in-context information, RAG provides a more modular and cost-efficient framework for knowledge management. Specifically, RAG excels in situations requiring access to dynamically updated or domain-specific knowledge bases, where external retrieval ensures factual freshness without retraining the model. It also reduces computational cost, as only the relevant subset of information is retrieved and processed, rather than encoding the entire knowledge corpus into the prompt. Thus, RAG remains particularly advantageous in enterprise, research, and real-time information retrieval settings where scalability, controllability, and verifiability are crucial.

\textbf{Open Issues.}
Despite these strengths, long-context LLMs demonstrate competitive performance in tasks requiring holistic reasoning over continuous text, such as narrative understanding or legal document analysis, where retrieval boundaries may fragment context. Empirical studies have also shown that model performance tends to degrade for tokens located in the middle or far regions of extremely long contexts \cite{liu-etal-2024-lost}, suggesting that larger windows do not guarantee better reasoning. The two paradigms are not mutually exclusive; hybrid architectures that integrate RAG’s retrieval precision with long-context reasoning are emerging as a promising direction. For example, adaptive retrieval strategies can selectively extend the model’s context window based on query complexity, while hierarchical memory frameworks can combine parametric, retrieved, and in-context knowledge. Future research should focus on quantifying the cost-performance trade-offs between these paradigms, identifying task-specific thresholds where RAG or long-context inference becomes more efficient. Such exploration will be crucial for determining the next generation of scalable, context-aware knowledge systems.

\subsection{Developments of Benchmark and Evaluation} \textbf{Significance.} With the rapid evolution of RAG techniques, there is a growing need for robust evaluation methodologies and standardized benchmark datasets. Comprehensive, reproducible, and challenging benchmarks are crucial for making objective comparisons among different approaches and for identifying each method’s strengths and weaknesses. In real-world settings, where multimodal inputs, personalization, and online scenarios are increasingly common, it is necessary to balance and expand metrics encompassing accuracy, efficiency, robustness, and user experience.

\textbf{Open issues.} Existing benchmarks often focus on single-modality or offline use cases, making them insufficient to capture the diverse application requirements of future RAG deployments. Moreover, there is still no unified standard for quantifying model interpretability, trustworthiness, or uncertainty handling. As online services and privacy compliance become prevalent, evaluation methodologies also need to address real-time constraints and security considerations to provide more comprehensive experimental foundations for new RAG research and applications.

\section{Conclusion}
In this paper, we have conducted a systematic and extensive review of the most notable works to date on RAG from a knowledge-centric perspective. We proposed an overarching framework that organizes existing research around the lifecycle of knowledge within RAG systems, from sourcing and retrieval to integration and generation, to clarify the relationships and trade-offs between different methods. By analyzing the challenges present at each stage of this lifecycle, particularly in knowledge parsing, integration, and context adaptation, we offered a classification scheme that contextualizes emerging approaches such as multimodal and memory-augmented RAG. We also identified critical open issues and promising research directions, underscoring RAG’s potential to transform knowledge-intensive applications across diverse domains. We hope that this survey will provide researchers with a solid understanding of the key components, main challenges, and ongoing developments in RAG, and shed light on new opportunities in future studies.

\section*{Acknowledgments}
This work was supported by grants from the National Key Research and Development Program of China (Grant No. 2024YFC3308200), the National Natural Science Foundation of China (62525606,62502486), the Fundamental Research Funds for the Central Universities of China (No. WK2150110032), USTC Kunpeng-Ascend Scientific and Education Innovation Excellence Center.



\bibliographystyle{ACM-Reference-Format}
\bibliography{samples/survey_rag}

\appendix

\end{document}